\pdfoutput=1
\documentclass{article}
\usepackage{arxiv}
\usepackage[utf8]{inputenc}
\usepackage[T1]{fontenc}
\usepackage{url}
\usepackage{booktabs}
\usepackage{amsfonts}
\usepackage{nicefrac}
\usepackage{microtype}
\usepackage{lipsum}

\usepackage{longtable}
\usepackage{tabularx}
\usepackage{xcolor}
\usepackage{tabulary}
\usepackage{multirow}
\usepackage{graphicx}
\usepackage{caption}
\usepackage{subcaption}
\usepackage{titlesec} 
\usepackage[numbers]{natbib}

\title{Explainable Goal-driven Agents and Robots - A Comprehensive Review}

\author{
  Fatai Sado, Chu Kiong Loo\thanks{Corresponding author \newline \emph{This research was supported by the Georg Forster Research Fellowship for Experienced Researchers from Alexander von{ Humboldt-Stiftung/Foundation and IIRG Grant (IIRG002C-19HWB) from University of Malaya}}} , Wei Shiung Liew \\
  Department of Artificial Intelligence\\ Faculty of Computer Science and Information Technology\\ University of Malaya \\ 
  50603 Kuala Lumpur, Malaysia \\ \texttt{abdfsado1@gmail.com, ckloo.um@um.edu.my, liew.wei.shiung@siswa.um.edu.my}\\
 \And
  Matthias Kerzel, Stefan Wermter \\
  Department of Informatics\\ Knowledge Technology\\ University of Hamburg\\
  Vogt-Koelln-Strasse 30 \\ 22527 Hamburg, Germany \\
  \texttt{matthias.kerzel@uni-hamburg.de, stefan.wermter@uni-hamburg.de} \\

}

\begin{document}
\maketitle

\begin{abstract}
Recent applications of autonomous agents and robots have brought attention to crucial trust-related challenges associated with the current generation of artificial intelligence (AI) systems. AI systems based on the connectionist deep learning neural network approach lack capabilities of explaining their decisions and actions to others, despite their great successes. Without symbolic interpretation capabilities, they are 'black boxes', which renders their choices or actions opaque, making it difficult to trust them in safety-critical applications. The recent stance on the explainability of AI systems has witnessed several approaches to eXplainable Artificial Intelligence (XAI); however, most of the studies have focused on data-driven XAI systems applied in computational sciences. Studies addressing the increasingly pervasive goal-driven agents and robots are sparse at this point in time. This paper reviews approaches on explainable goal-driven intelligent agents and robots, focusing on techniques for explaining and communicating agents' perceptual functions (e.g., senses, vision) and cognitive reasoning (e.g., beliefs, desires, intention, plans, and goals) with humans in the loop. The review highlights key strategies that emphasize transparency, understandability, and continual learning for explainability. Finally, the paper presents requirements for explainability and suggests a road map for the possible realization of effective goal-driven explainable agents and robots.
\end{abstract}


\keywords{Accountability, continual learning, deep neural network, explainability, explainable AI, goal-driven agents, transparency}

\maketitle

\section{Introduction}

\subsection{Background/Motivation}

One of the goals of human-robot partnership is to establish teams of humans and robots that effectively make use of the individual capabilities of each team member \cite{scholtz2003theory}. Robots are used in situations involving boring and repetitive work or for tasks that involve a significant risk of injury. Humans, on the other hand, play the role of supervisor, operator, mechanic, bystander, or teammate \cite{scholtz2003theory}. Having more than one robot member being supervised by one human should improve the efficiency and productivity of the team. Hence, giving the robots more autonomy and the ability to act semi-independently would give the human more freedom to supervise multiple robots simultaneously.

Goal-driven artificial intelligences (GDAIs) include agents and robots that are autonomous, capable of interacting independently within their environment to accomplish some given or self-generated goals \cite{biran2017explanation}. These agents should possess human-like learning capabilities such as perception (e.g., sensory input, user input) and cognition (e.g., learning, planning, beliefs). GDAIs engage in tasks that require activity over time, generate plans or goals, and execute their ideas in the environment, applying both perception and cognition. They can also adapt the intentions or purposes as the need arises and may be required to account for their actions \cite{lake2017building} \cite{carey2018data} and embody such capabilities in the context of lifelong developmental learning \cite{wermter2017crossmodal}. GDAIs are helpful in several applications including exploration in hazardous areas, transportation, and gaming \cite{carey2018data}.

In addition to problem-solving skills, the agents should provide a causal explanation for their decisions and reasoning. A relevant example in this context involves an autonomous robot that plans and carries out an exploration task. It then participates in a debriefing session to provide a summary report and addresses a human supervisor's questions. Agents must explain the decisions they made during plan generation, stating considered alternatives, to report which actions they executed and why, explain how actual events diverged from the plan and how they adapted in response, and communicate decisions and reasons in a human-understandable way to gain the user's trust \cite{choo2018visual} \cite{baum2018machine} \cite{swartout1993explanation}.

This review focuses on two aspects of human-like learning for GDAIs: explainability and continual learning of explanatory knowledge. We focus on both situated/non-situated and embodied/non-embodied autonomous eXplainable GDAIs (XGDAIs). "Situated agents" are located within the environment that they can perceive and operate in, as opposed to "non-situated agents" which are located in a remote location. "Embodied agents" have a physical form  while "non-embodied agents" are purely digital.

The review categorizes explanation generation techniques for XGDAIs according to the agent's behavioral attributes, i.e., reactive, deliberative, and hybrid. It provides a clear taxonomy of XGDAIs and clarifies the notion of the agent's behavioral attributes that influence explainability. For each category, the review focuses on explanation at the level of the agent's perception (e.g., sensory skills, vision) and cognition (e.g., plans, goals, beliefs,desires, and intentions). While an agent's perceptual foundation (dominated by reactive reasoning) may be connected to the sub-symbolic reasoning part relating the agent's states, vision, or sensors/environmental information to the agent's cognitive base, the cognitive base (dominated by deliberative reasoning) relates plans, goals, beliefs, or desires to executed actions. Finally, we provide a road-map recommendation for the effective actualization of XGDAIs with an extended perceptual and cognitive explanation capability.

\subsection{What Is eXplainable AI And Why Is Explainability Needed?}

XAI refers to machine-learning or artificial intelligence systems capable of explaining their behavior in human-understandable ways \cite{ehsan2019automated}. Explanations help humans collaborate with an autonomous or semi-autonomous agent to understand why the agent fails to achieve a goal or unexpectedly completes a task. For instance, if an autonomous agent fails to finish a task or performs an unexpected action, the human collaborator would naturally be curious to understand how and why it happens. Explanations thus enable the human collaborator to comprehend the dynamics leading to the agent's actions, enabling the human to decide how to deal with that behavior.

Although the need for explainability in AI systems has been long established during the MYCIN era, also known as the era of the expert systems \cite{preece2018stakeholders} \cite{puppe2012systematic}, the current drive for explainability has been motivated by recent governmental efforts from the European Union, United States (USA) \cite{bundy2017preparing}, and China \cite{mittelstadt2017transparent}, which have identified AI and robotics as economic priorities. The European Union's key recommendation is the right to explanation which is stressed by the General Data Protection Regulation (GDPR) \cite{carey2018data} \cite{anjomshoae2019explainable}. AI systems must explain their decisions, actions, or predictions for safety-critical applications to ensure transparency, explainability, and accountability \cite{mittelstadt2017transparent}.

\subsection{What Is Data-driven And Goal-driven XAI?}

In machine learning, explainability in data-driven AI is often linked to interpretability. According to \citet{biran2017explanation}, a particular system is interpretable if a human can understand its operations through explanation or introspection. \citet{choo2018visual} describe a deep learning neural network's (DNN) interpretability as determining the input attributes that account for output predictions. Thus, data-driven XAI implies explaining the decision made by a machine learning "black-box" system, given input data \cite{guidotti2018survey}. The motivation to discover how available data contributes to a decision is an important aspect of this branch of XAI and whether the machine learning process can reliably replicate the same decision, given similar data and specific circumstances \cite{holzinger2017towards}. 

On the other hand, goal-driven XAI is a research domain that aims to create explainable robots or agents that can justify their behaviors to a lay user \cite{anjomshoae2019explainable}. The explanations would assist the user in creating a Theory of Mind (ToM), comprehending the agent's behavior, and contribute to greater collaboration between the user and the agent. A better ToM of the agent would enable the user to understand the agent's limitations and capabilities, thus enhancing confidence and safety levels and preventing failures. The absence of adequate mental models and understanding of the agent can contribute to a failure of interactions \cite{anjomshoae2019explainable} \cite{chandrasekaran2017takes}. \citet{hase2020evaluating} study how techniques used for generating explanations affect the users' ability to predict what the explaining agent will do in a given situation. In addition, they discover that what a user considers a good explanation is not the same as how helpful the explanations are. As discussed in the next subsection, there is a growing application of goal-driven XAI in the current AI-dependent world \cite{langley2017explainable}.

\subsection{Emerging Trends In XGDAI}

\begin{figure}[!b]
    \centering
    \includegraphics[width=0.9\linewidth]{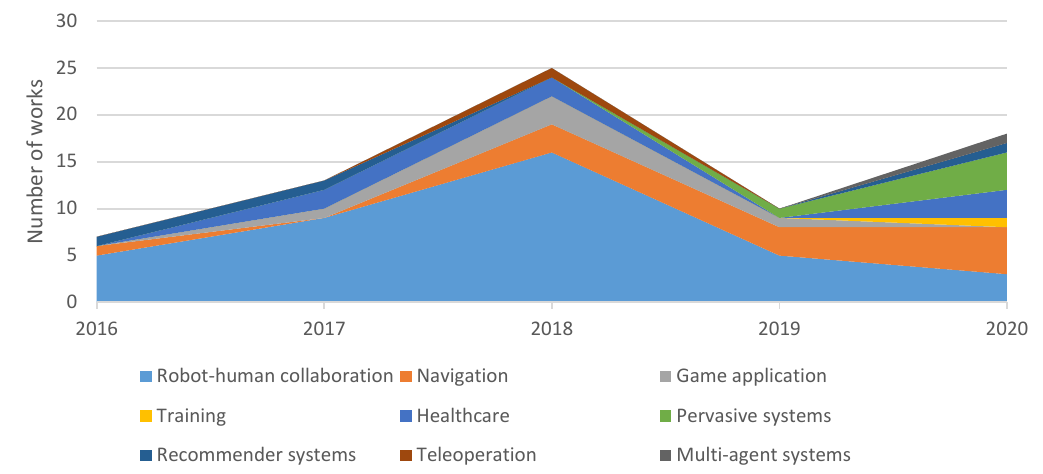}
    \caption{Five-year trend of the use of explainable techniques in selected application domains.}
    \label{fig_graphtrend}
\end{figure}

Fig. \ref{fig_graphtrend} shows the distribution of works on XGDAIs from 2016 to 2020 by selected application domains. Literature search for relevant works was performed using the search engines Google Scholar and Science Direct. Combinations of search keywords were used using Boolean operators and quote marks to locate the relevant terminology. For example, ("artificial intelligence") AND ("explainable" OR "transparent" OR "legibility" OR "interpretability") AND ("agent" OR "robot") AND ("goal-driven"). The search results were further filtered by examining the text of the abstracts to fulfil the following criteria: 

\begin{enumerate}
    \item The paper was published between 2016 to 2020 and the full text must be available online.
    \item The paper must be relevant to the explainable AI domain.
    \item The paper must be a primary study, i.e. introducing a novel model, method, or architecture. Survey and review papers were excluded.
    \item The paper must involve the use of robots, agents, or AI systems to fulfil a specific function, i.e. "goal-driven". The terminology is not a requirement, however. Several works that fulfil this criterion but do not explicitly call it "goal-driven" were included.
    \item The paper must include explainability methods that are explicitly communicated to the user. Methods that fulfil this criterion include natural language explanations, visualizations, etc.
\end{enumerate}

In 2016, research into XGDAIs focused on human-robot collaborative tasks, navigation, and healthcare. However, studies in XGDAIs branched out to other domains over the next few years. This upsurge can be seen as the effect of the public pressure on the explainability of AI systems and initiatives by several national government agencies like the "right to explanation" by the GDPR \cite{carey2018data} \cite{anjomshoae2019explainable}. This trend is likely to continue as AI adoption progresses in other fields. Research into explainability for pervasive systems, for example, can be attributed to the growth of increasingly "smart" and complex systems for both domestic and industrial use. Research into navigation systems is also growing alongside interest in self-driving vehicles. The studies are shown in Table \ref{tab1_supp} in detail, along with significant taxonomies.

\section{Terminology Classification}

\subsection{Attributes Of XGDAI}

The studies obtained from the literature search as described in the previous section were taxonomized based on several categories: 

\begin{enumerate}
    \item Behavioral architecture: whether the architecture of the model was deliberative, reactive, or hybrid.
    \item Application domain: what the model was used for.
    \item Transparency: whether the internal workings of the model were observable or explainable.
    \item Domain specificity: whether the model can be used for only one domain or was domain-agnostic.
\end{enumerate}

Regarding the agent's behavior and interaction with the world, three behavioral architectures are traditionally distinguished in literature for XGDAI: deliberative – in which the agent deliberates on its goals, plans, or actions, or acts based on a sense-model-plan-act cycle (the agent, in this case, should possess a symbolic representation of the world); reactive – in which the agent implements some simple behavioral patterns and reacts to activities or events in its environment in a stimulus-response way, no model of the world is required (the robot chooses one action at a time), and hybrid - which combines the above two behaviors \cite{davidsson1996autonomous}. Each of the three behavioral architectures can be further sub-divided based on the explainability method used, such as goal-driven autonomy, goal-driven agency, Belief-Desire-Intention (BDI), and so on. Table \ref{tab1_supp} in the supplementary materials presents the taxonomies of XGDAI behavior found in the literature. In this section, we make further clarification of these attributes.

\subsubsection*{Deliberative} Deliberative agents act more like they think, searching through a behavioral space, keeping an internal state, and predicting the consequences of an action. \citet{wooldridge1995conceptualising} describes such an agent as one with a symbolic model of its world that could make decisions based on symbolic reasoning or rationale. According to the conventional approach, the agents' cognitive aspect consists of two parts: a planner and a world model \cite{davidsson1996autonomous}. The world model is an internal representation of the agent itself in the external environment. The planner model uses this representation to make a strategy for how the agent's goals can be accomplished. How such agents operate may be viewed as a 'sense-model-plan-act' behavior. The BDI model is the most widely used architecture to enforce such actions, where the beliefs of an agent about the world (its picture of the world), desires (goal), and intentions are internally expressed, and realistic reasoning is applied to select an action \cite{hayzelden1999software}.

\subsubsection*{Reactive} Reactive agents present a collection of simple behavioral patterns that react to environmental changes \cite{davidsson1996autonomous}, no model of the world is included. They achieve their goal by reacting reflexively to external stimuli, choosing one action at a time. The creation of purely reactive agents came at the heels of the limitations of symbolic AI. Developers of reactive agent architectures rejected symbolic representations and manipulation as a base of AI \cite{knight1993many}. Model-free (deep) reinforcement learning (RL) is a state-of-the-art approach that enables reactive agent behavior. Some notable explainability works in RL include Memory-based eXplainable Reinforcement Learning (MXRL) \cite{cruz2019memory}, Minimal Sufficient eXplanation (MSX) via Reward Decomposition \cite{juozapaitis2019explainable}, and Reward Augmentation and Repair through Explanation (RARE) \cite{tabrez2019improving}. Previous reviews on reactive RL agents can be found in \citet{lin1992self}. 

\subsubsection*{Hybrid} A significant number of researches have focused on combining reactive and deliberative agent techniques, leading to creating a compound called a hybrid agent, which combines comprehensive internal manipulation with non-trivial symbolic constructs and external events with reflexive reactive responses \cite{davidsson1996autonomous}. The modern drive integrates reactivity's flexibility and robustness with deliberation's foresight \cite{korpan2018toward}. Supporters of the hybrid approach believe it is optimal since high-level deliberative reasoning and low-level response to perceptual stimuli seem essential for actualizing an agent's behavior. An example is the hybrid system proposed by \citet{wang2008robot}. Reactive exploration is used to create waypoints, which are then used by a deliberative method to plan future movements in the same area. Another existing system which mixes reactive and deliberative behaviors is the agent developed by \citet{rao1995bdi}, which determines when to be reactive and when to pursue goal-driven plans.

\subsection{Application Scenarios For XGDAI}

This section presents a number of application scenarios for XGDAI that are primarily reported in the literature. These are summarized in Table \ref{tab1_supp} in the supplementary materials.

\begin{itemize}
    \item \textbf{Robot-human collaborative tasks} involve close interactions with humans in factory settings, and teaming in an outdoor setting are the predominantly mentioned application scenarios in literature. In robot-human collaborative scenarios, explainability (i.e., transparency) of XGDAI was shown to enhance the quality of task performance \cite{breazeal2005effects} and to enable both robots and humans to take responsibility (credit or blame) for their actions in collaborative situations \cite{kim2006should}.
    
    \item \textbf{Robot navigation}, where the robot or agent-controlled vehicle moves through an environment towards a goal while avoiding obstacles. Examples of navigation applications include solving a maze or self-driving vehicles. Explanations are generated in this domain to explain why specific routes are chosen over others or what obstacles are detected and avoided.
    
    \item \textbf{Game applications} used explainability techniques to describe gameplay behavior, whether of non-player characters (NPCs) \cite{molineaux2018towards} or player characters \cite{madumal2019explainable}. The latter is used to decompose the behavior of human-controlled players into discrete elements that can then be programmed into computer-controlled agents to emulate human gameplay.

    \item \textbf{Healthcare} applications employed agents and robots for medical diagnosis and as caregivers. Autonomous agents in this field are strictly regulated as any error may potentially harm a patient. Using explainer systems, the agents' steps to arrive at a diagnosis can be traced. Medical experts acting as "human-in-the-loop" can observe the agents' justifications and correct them if an erroneous diagnosis is performed.
    
    \item \textbf{Training} or teaching scenarios involve agents in the role of either instructors or students. As instructors, the agents analyze students' strengths and weaknesses to create individualized teaching syllabi. As students, agents emulate human behavior of, for example, pilots in cockpits. Explanation generation methods are used for the latter to diagnose their reasoning and decisions.

    \item \textbf{Pervasive systems} or ubiquitous computing utilized agents to manage highly complex systems and monitor many variables simultaneously. Explainability simplifies the system operations, making it easier and quicker for human operators to intervene and diagnose errors \cite{vermeulen2010improving}.

    \item \textbf{Recommender systems} are typically user interfaces to match a user's preferences to a large database of products. Given a large number of user variables, the system searches for the closest products that fulfill all user requirements. Recommenders with explainability systems can provide a larger selection of products and explain how each product matches the user's preferences.

    \item \textbf{Multi-agent systems} relate to using multiple agents cooperating to accomplish a goal. Emergent behavior occurs when interactions of multiple agents produce outcomes unforeseen by the programmers. Explainer systems provide insights on how and why individual agents take certain actions and how the cumulative choices produce emergent behavior \cite{ho2020explainable}.
\end{itemize}

\subsection{Transparency}
Transparency refers to whether the internal workings of the XGDAI are readily observable. Observable processes should include from the moment the model receives sensory input to when the model produces the final output decision, as well as every intermediate step. In addition, the observable elements should be in a format that is easily interpretable, for example using natural language to describe the processes or visualization techniques to highlight significant areas. Transparent models by their nature should be replicable; a human user should be able to explain how and why a specific input arrives at a specific output.

\section{Explanation Generation Techniques for XGDAI}

\begin{figure}[!tb]
    \centering
    \includegraphics[width=0.8\linewidth]{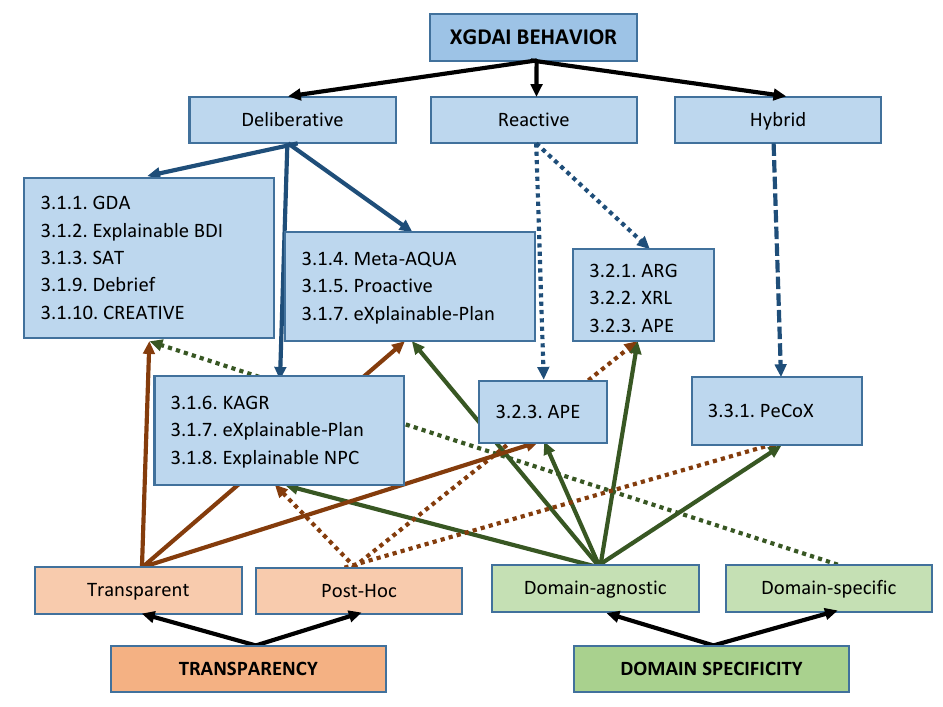}
    \caption{Taxonomy of the reviewed XGDAI.}
    \label{fig_taxonomy_map}
\end{figure}

This section presents existing explanation generation techniques and taxonomies (e.g., transparency, domain dependence, post-hoc explainability, continual learning) for XGDAI. The section is subdivided into three parts to discuss techniques categorized under Deliberative, Reactive, or Hybrid Explanation Generation. The techniques can be further categorized by Transparency (Transparent vs Post-Hoc) and Domain Specificity (Specific vs Agnostic). Domain-specific techniques heavily depend on the agent world's domain knowledge and do not permit application to other agents in other environments. Domain-agnostic techniques are domain-independent, allowing cross-platform usability. Post-hoc explanations make it possible to explain without inherently following the reasoning process leading to the decision \cite{anjomshoae2019explainable}. Transparent explanations on the other hand highlight every step of the reasoning process. Fig. \ref{fig_taxonomy_map} maps the reviewed XGDAI to three taxonomy categories: XGDAI Behavior, Transparency, and Domain Specificity.

\subsection{Deliberative XGDAI}
From the literature search, models are classified as Deliberative if they have two components: an internal 
representation of the external environment, and a planner to design a strategy to accomplish the agent's goals with 
respect to the environment. 

\subsubsection{Models of Goal-driven Autonomy - Transparent domain-agnostic}

\begin{figure}[!htb]
\centering
\includegraphics[width=0.6\textwidth]{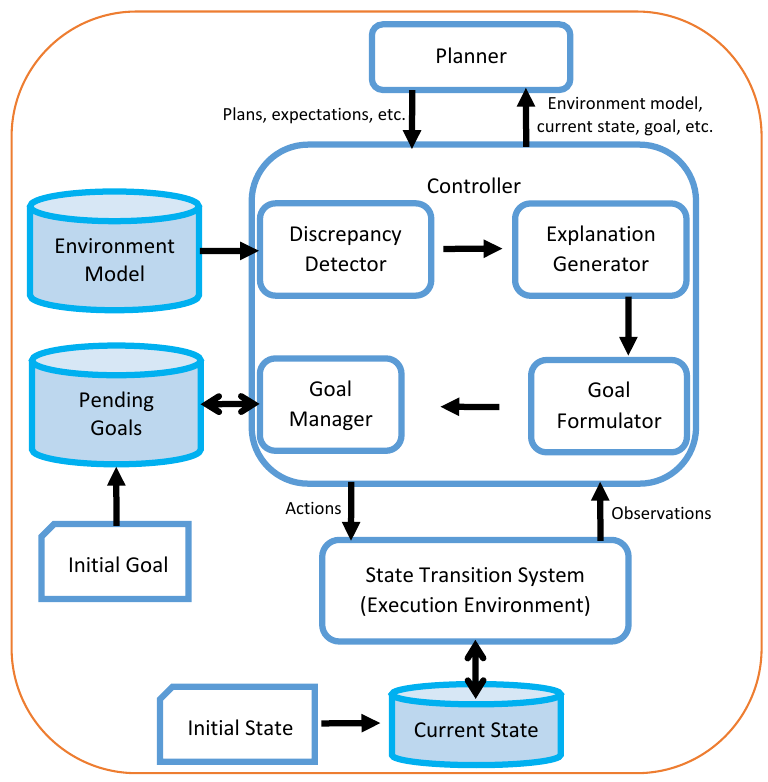}
\captionsetup{justification=raggedright}
\caption{The ARTUE goal-driven autonomy conceptual model \cite{molineaux2010goal}. The model extends the \citet{nau2007current} online planning model. The controller communicates with the execution environment (i.e., state transition system) and the planner. As input, the planner receives a planning problem consisting of an environmental model, the current state of the system, and a potential goal. The planner generates a series of action plans and expectations (states constraints when executing actions). The controller sends the action plan to the state transition system, then processes the subsequent observations. The controller detects discrepancies, produces explanations, generates goals, and manages the goals.}
\label{fig02}
\end{figure}

Goal-driven autonomy is a conceptual goal justification mechanism that enables an agent to generate explanations when detecting a mismatch in the environment state. The model enables the robot to continuously track the execution of its current plan, determine whether the states met expectations, and generate an explanation upon identifying a mismatch \cite{molineaux2010goal}. The model is also extended to enable an agent to identify when new goals should be chosen and explain why new goals should be followed \cite{jaidee2011integrated}.

In many existing works in goal-driven autonomy, explanations are generated when active goals are completed or when discrepancies are detected between the actual event and the intended plan. Differences may arise due to one of several factors; if the domain knowledge is flawed, if the perception of a state is inaccurate, or if there is an unseen factor influencing the state. An explanation is generated to explain the discrepancies and find or address the hidden factors \cite{molineaux2010goal}. 

The Autonomous Response To Unexpected Events (ARTUE) \cite{molineaux2010goal} is a domain-independent goal-driven agent that continuously reacts to unexpected situations by reasoning on what goals to pursue to resolve the situation. ARTUE is designed to deal with unexpected environmental situations by first explaining these changes and then developing new goals which integrate the explicit information on the hidden environmental aspects. On this basis, ARTUE should deal with new and unobserved objects in the planning. The ARTUE system is implemented in a sandbox simulation scenario \cite{auslander2009towards}. The ARTUE system as shown in Fig. \ref{fig02} integrates a planner that deliberates on exogenous events through predicting future states in a dynamic environment, and a controller that identifies discrepancies, generates explanations, generates targets, and manages goals. The explanation aspect deliberates on the world's hidden knowledge by abductive reasoning on the conditions and effects of the planning \cite{reiter1986assumption}.

\subsubsection{Explainable BDI Model - Transparent domain-specific}

BDI agents or robots, primarily symbolic AIs with integrated beliefs, desires, and intentions, offer good prospects for producing useful and human-understandable explanations \cite{harbers2010design}. According to \citet{bratman1987intention}, belief and desire are mental attitudes (pro-attitudes) driving action. Intention is distinguished as conduct controlling this pro-attitude, which can be treated as elements of partial plans of action \cite{bratman1987intention} \cite{georgeff1998belief}. \citet{harbers2010design} propose a BDI model that allows the explanation of a BDI agent's actions (behavior) based on the underlying beliefs and goals. Goals are described as "active" desires that the agent currently pursues. An example of an explanation of actions based on 'belief' can be: "The boy opened the window because he believed someone was outside", and based on 'goals' can be: "The boy opened the window because he wanted to see who was outside". The motivation here is that humans can clarify and comprehend their behaviors or actions in terms of underlying beliefs, desires, intentions, or goals \cite{keil2006explanation} \cite{malle1999people}. Thus, since BDI agents establish their actions by deliberation on their mental state, the mental principles behind the action are applied to interpret their actions. Also, since mental reasoning is expressed symbolically, the explanation generation process is straightforward. Typically, a log of behavior preserves all the agent's prior mental states and actions that could be used for explanations. On request, the BDI algorithm is implemented on the log and selects the beliefs and goals that become part of the explanation. However, not all 'explanatory elements' can be helpful in the explanation \cite{keil2006explanation}.

An important aspect of explainable BDI agents is that they can clarify typical human errors. According to \citet{flin2017incident}, revealing explainable BDI agents' actual mental state to trainees may make them aware of their (false) assumptions about them. People may make false assumptions about others' experiences and motives in many crucial circumstances \cite{flin2017incident}, a phenomenon of attributing incorrect mental states to others \cite{keysar2003limits}.

\subsubsection{Situation Awareness–based Agent Transparency (SAT) Model - Transparent domain-specific}

SAT is an agent transparency model for enhancing an operator's situational awareness \cite{endsley2018innovative} in the agent's environment \cite{chen2014situation} \cite{boyce2015effects}. The SAT depends on the BDI model framework and is implemented as a user interface to provide transparency to the operator. An important extension is that it provides transparency not only of the status of the robot (e.g., plans, current state, and goals) and process of reasoning but also of the future projections of the robot (e.g., future environment states) \cite{chen2018situation}. Basic information on the robot's current goal and state, intentions, and expected action is presented to the operator at the entry phase of the SAT structure. In the second stage, knowledge about the agent's reasoning process supporting its action and the limitations it takes into consideration are presented before the operator. In the third stage, specifics of the robot's future projection are given to the operator, such as expected outcomes, the possibility of failure or success, and any ambiguity inherent in the predictions. The transparency of an agent in this context is an informative ability to provide an operator's understanding of the purpose, results, plans, reasoning process, and future projections of an agent. The operator's trust is described as the agent's willingness to help achieve the operator's goals in especially uncertain and vulnerable circumstances \cite{anjomshoae2019explainable}. SAT requires highly specific information with regards to the agent, task, and environment combined to create a dynamic construct. As each new input arrives, the SAT incorporates both temporal ("what happened when") and spatial information ("what happened where"). The internal model integrates various information and understands their context relevant to goals in order to come up with appropriate actions for the situation. Due to the significant effort required for designing the SAT model to fit the application, the method was considered domain-specific.

\subsubsection{Meta-AQUA Explanation Model - Transparent domain-agnostic}

Meta-AQUA is an introspective learning system proposed by \citet{cox2007perpetual} for a self-aware cognitive agent. The system can allow an agent to decide (learn) its goals by understanding and describing unusual occurrences in the world. The learned objectives seek to modify the agents' perception of the world to minimize the dissonance between what the agent expects and the world's actual reality. With intelligent behaviors, Meta-AQUA integrates cognitive components such as planning, understanding, and learning and metacognition such as cognition monitoring and control. Learning in this context is described as a deliberate planning task with learning goals.

In contrast, the explanation of unusual events is a key to enhancing the agent's learning of a new goal. Meta-AQUA uses its meta-reasoning component (i.e., meta-cognition) to explain a reasoning or expectation failure, enhancing its goal learning. It adopts case-based knowledge representations as frames linked together by explanatory patterns to represent general causal structures \cite{cox2007perpetual} \cite{cox1999introspective}.

\begin{figure}[!htb]
\centering
\includegraphics[width=0.6\textwidth]{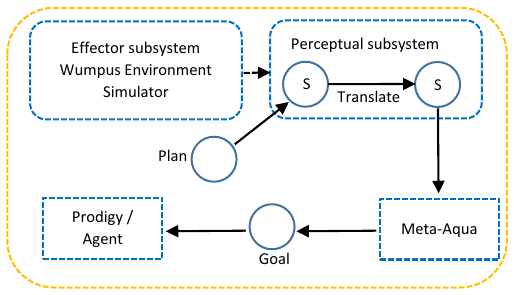}
\captionsetup{justification=raggedright}
\caption{The INTRO architecture \cite{cox2007perpetual} consists of a primitive subsystem of perceptions and effectors and two components of cognitive planning and comprehension.}
\label{fig03}
\end{figure}

Meta-AQUA is implemented in the INitial inTROspective cognitive agent (INTRO) and simulated in the Wumpus World-simulated environment, a partially observable environment where the objective of the agent or robot is to seek a pot of gold while evading pits and the Wumpus creature (Fig. \ref{fig03}). INTRO is designed with primitive perceptual and effector subsystems and two cognitive components.

INTRO has four main components: a primitive subsystem of perceptions and effectors and two components of cognitive planning and comprehension. The cognitive components compose of the respective prodigy/agent and Meta-AQUA structures. The perceptual subsystem acts as a translator for the cognitive subsystems. The central cognitive cycle is to explore the world, form an objective of changing the world by understanding and describing unusual events or world states, establish a plan to accomplish the goal, and finally perform according to the plan, observing the outcomes. Instead of using explanation to alter its understanding, INTRO uses explanation to create a goal to improve the environment.

The agent will observe the world with these components, develop a goal to transform the world, devise a strategy for achieving the objective, and eventually change the environment, including turning, going forward, picking up, and firing an arrow. As the agent navigates the environment, the Meta-AQUA system seeks to explain the effects of the agent's behavior by inserting the events into a conceptual representation and constructs an internal model to reflect the causal ties between them. Meta-AQUA creates an anomaly or other interesting occurrence to produce an explanation of the event. However, a major aspect of this approach is that the agent utilizes the explanation to produce a goal to modify the environment (e.g., shoot an arrow) instead of using the explanation to change its understanding.

\subsubsection{Proactive Explanation Model - Transparent domain-agnostic}

In scenarios that involve teams of humans and autonomous agents, a proactive explanation that anticipates potential surprises can be useful. By providing timely explanations that prevent surprise, autonomous agents could avoid perceived faulty behavior and other trust-related issues to collaborate effectively with team members. This is an important extension to explainability for agents since most techniques are usually provided to respond to users' queries. In this context, \citet{gervasio2018explanation} present a conceptual framework that proposes explanations to avert surprise, given a potential expectation violation. The proactive explanation attempts to reconcile the human mental model with the agent's real decision-making process to minimize surprise that could disrupt the team's collaboration. Surprise is the primary motivation for pro-activity and is triggered if the agent's action deviates from its past actions, or if its action is considered unusual or contrary to plan \cite{gervasio2018explanation}.

\subsubsection{KAGR Explanation Model – Post-hoc domain-agnostic}

\begin{figure}[b!]
\centering
\includegraphics[width=0.6\textwidth]{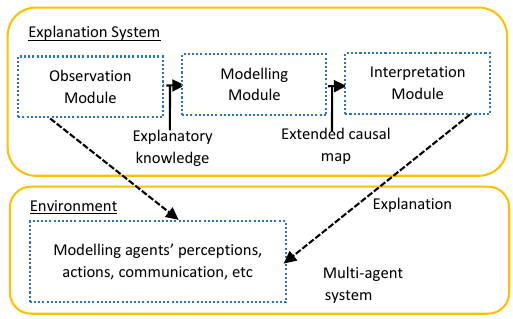}
\captionsetup{justification=raggedright}
\caption{KAGR explanation architecture \cite{sbai2013intra}. The system consists of an observation module for the generation of the explanatory knowledge, a modeling module for the knowledge formalism using an extended causal map (CM), and an interpretation module to produce explanations.}
\label{fig04}
\end{figure}

KAGR is an explanation structure proposed by \citet{sbai2013intra} to explain agent reasoning in multi-agent systems (MAS) (operating in uncontrollable, dynamic, or complex situations) where an agent's reasoning is not reproducible for the users. KAGR describes an agent's reasoning state at run-time as a quadruplet <K,A,G,R> (Knowledge, Action, Goal, Relation), also considered explanatory knowledge. Under the KAGR conceptual explanation framework, events produced by the agents at run-time are intercepted, and an explanatory knowledge acquisition phase is performed to represent the knowledge attributes or details related to the execution of events in a KAGR structure \cite{sbai2013intra} \cite{hedhili2013explanation}. A further step is performed to generate a knowledge representation formalism by linking the attributes in an extended CM model. A final step achieves natural language interpretation for the CM model using predicate first-order logic to build up a knowledge-based system for an understandable reasoning explanation to users. Overall, as seen in Fig. \ref{fig04}, it adopts a three-module architecture: a first module that generates explanation, another module that puts the knowledge in the formalism of the extended causal chart, and a third module that uses first-order logic to analyze and interpret the constructed CM to produce reasonable explanations.

\subsubsection{eXplainable Plan Models – Post-hoc/Transparent domain-agnostic}

An aspect of planning is Plan Explanation, in which the primary objective is to enable individuals to understand the plans of the (agent) planner (e.g., \cite{sohrabi2011preferred} \cite{seegebarth2012making} \cite{bidot2010verbal}), including interpreting the agents' plans into human-understandable form and the creation of interfaces that facilitate this understanding. In this context, related works include XAI-Plan \cite{borgo2018towards}, Refinement-Based Planning (RBP) \cite{bidot2010verbal}, and WHY-Plan \cite{korpan2018toward}.

\begin{figure}[b!]
\centering
\includegraphics[width=0.7\textwidth]{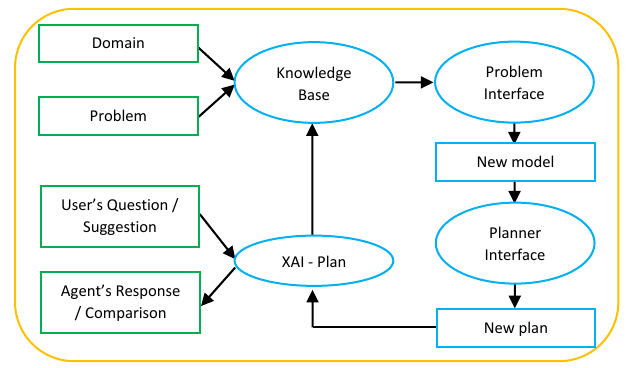}
\captionsetup{justification=raggedright}
\caption{Example XAI–Plan architecture by \citet{borgo2018towards}. The XAI-Plan node generates explanatory plans and communicates agent's response through a user interface. ROSPlan provides the planner interface, problem interface, and knowledge base.}
\label{fig05}
\end{figure}

XAI-Plan is an explanatory plan model proposed by \citet{borgo2018towards} to provide an immediate explanation for the decisions taken by the agent planner. The model produces explanations by encouraging users to try different alternatives in the plans and compares the subsequent plans with those of the planner. The interactions between the planner and the user enhance hybrid-initiative planning that can improve the final plan. The XAI-Plan planning system is domain-agnostic and independent. XAI-Plan answers questions such as "why does the plan include a specific action and not another similar action?". The algorithm uses a preliminary set of plans as input; the user chooses one of the plan's actions; the XAI-Plan node generates explanatory plans and communicates with the user through the user interface (Fig. \ref{fig05}). ROSPlan provides the planner interface, problem interface, and knowledge base, which is used to store a Planning Domain Definition Language (PDDL) model and provide the AI planner (i.e., an architecture for embedding task planning into ROS systems) with an interface.

RBP is a transparent domain-independent framework proposed by \citet{bidot2010verbal} to allow verbal queries from users and produce plan explanations verbally. RBP enables a transparent description of the search space examined during the planning process, giving the possibility to explore the search space backward to search for the relevant flaws and plan modifications. RBP builds on a hybrid planning structure incorporating hierarchical planning and planning using partial-order causal-link using states and action primitives \cite{biundo2014abstract}. 

\citet{korpan2018toward} propose WHY-Plan as an explanation method that contrasts the viewpoints of an autonomous robot and a human while planning a navigation route. The core of its explanation is how the planners' objectives differ. WHY-Plan addresses the question "Why does your plan go this way?" and exploits differences between planning objectives to produce meaningful, human-friendly explanations in natural language. An example of a WHY-Plan natural language explanation for a robot, in a team with a person, navigating from one location to another and avoiding a crowded path could be, "Although a somewhat shorter route exists, I think mine is a lot less crowded". The WHY-Plan's response compares two objectives: "avoid crowds", its planning objective, and an alternative objective: "take the shortest path", which can be attributed to the human questioner or team member. WHY-Plan is implemented in SemaFORR, a cognitive-based hybrid robot controller \cite{epstein2015learning}. SemaFORR can make a decision using two sets of reactive advisors.

\subsubsection{Explainable Non-Player Characters (NPC) – Post-hoc domain-agnostic}

\begin{figure}[b!]
\centering
\includegraphics[width=0.8\textwidth]{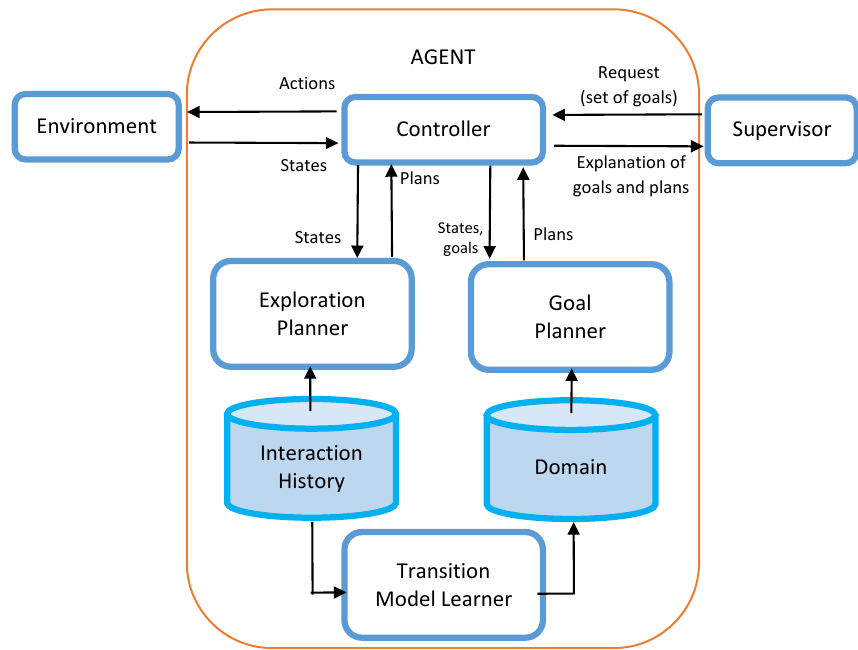}
\captionsetup{justification=raggedright}
\caption{Example of an Explainable NPC architecture \cite{molineaux2018towards}. The agent has four sub-modeled functionalities: the exploratory planner for obtaining new information; the goal planner necessary for implementing the supervisor's goals; the transition model learner responsible for updating the world model of the agent; and the controller responsible for determining when to explore.}
\label{fig06}
\end{figure}

Explainable NPC is an architecture proposed by \citet{molineaux2018towards} to minimize the frustration of video game players by explaining NPC actions. To many video game players, NPCs may become a source of frustration because their reasoning process is not transparent. The NPCs may respond to some internal necessity that a player is unaware of or may face barriers that a player cannot see, but they do not communicate these problems \cite{molineaux2018towards}. The NPC architecture is thus motivated to allow agents to know their environment, achieve goals and demonstrate to a supervisor what they are doing (Fig. \ref{fig06}). At each step, the architecture enables the agent to obtain an observation that represents knowledge about the environment's true state and to communicate with the environment (and supervisor) by taking action. The supervisor requests the agent that represents what the supervisor wants the agent to achieve, modified at each step. In response, the agent should always clarify why it takes a specific action to the supervisor. The framework is divided into four sub-modules: the exploratory planner, tasked with taking steps to collect new data to update an action model; the goal-directed planner, necessary for implementing the supervisor's goals; the learner of the transition model, necessary for updating the world model of the agent; and the controller, responsible for deciding when to explore \cite{molineaux2018towards}. 

In related work, \citet{van2004explainable} propose the XAI architecture for NPCs in a training system. The XAI functions to retrieve important events and decisions made from the replay log and enable NPCs to explain their conduct as an answer to the questions selected from the XAI menu. In this framework, each character's reactive behavior, low-level actions, and higher-level behavior generation are the responsibility of the NPC AI. The XAI system records the NPC AI's activities and utilizes the records or logs for post-action analyses during the execution process.

\subsubsection{Debrief – Transparent domain-specific}

Debrief is a multimedia explanation system proposed by \citet{johnson1994agents} that constructs explanations after an active goal. The circumstance in which the decision was taken was recalled and then replayed in the variants of the original situation, or by experimenting to determine which factors influenced the decision. The elements are considered critical as their absence would result in a different outcome of the decision process. The details of the agent's implementation, including individual rules and motivations applied in making a decision, are filtered out to construct an explanation. Under the framework, it is unnecessary to maintain a complete trace of rule firings to produce explanations. The system learns the relationships between situational factors and decisions to facilitate the explanation of similar decisions. Debrief is implemented in an artificial pilot simulator where the pilot can perform a flight patrol mission. After each task, the pilot is asked to describe and explain essential decisions it made along the way. The pilot has to clarify its evaluation of the situation. Debrief should enable the pilot to define the reasons for its actions and assess the situation and its beliefs that resulted in the actions.

\subsubsection{Explainable CREATIVE Model – Transparent domain-specific}

Cognitive Robot Equipped with Autonomous Tool InVention Expertise (CREATIVE) is a relational approach proposed by \citet{wicaksono2017towards}, enabling robots to learn how to use tools and design and construct new tools if needed. The critical information, such as learned observations, snapshots from a camera, and object positions, are stored in Prolog (Programming in Logic) to simplify the explanation of its tool creation process. CREATIVE utilizes relational representation, so its results have some inherent explainability features, establishing the relation between entities as facts in Prolog. A method of inductive logic programming (ILP) is employed to learn the relational representations of the tool models \cite{lavrac1994inductive}. A pre-determined set of basic questions and answers is used in the entire dialogue in Prolog.

\subsubsection{Summary on Deliberative XGDAIs}

Overall, reviewed literature on deliberative XGDAIs provides applicable explanation generation models and frameworks focusing on transparency and/or comprehensibility of agents' plans, goals, actions (behavior), reasoning processes, decisions, current states, and world states (i.e., unusual events and future projections of states). Leveraging on their model for plan execution or model of the world or tasks, their primary triggers or key basis for explanation generation include factors such as expectation failures (i.e., mismatches between what the agent expects and what the agent encounters), unusual events (in the world that is not previously modeled for the agent), anticipated surprise for the agent's action or decision, mismatched mental models (of the agent and human), users' requests, and an underlying belief or goal driving an action. These techniques assume certain aspects of the human-like explanation generation process. However, most approaches are domain-dependent, over-fitted to the particular situation or environment, limiting application to other domains.

Another concern for deliberative XGDAIs is how explanatory knowledge is acquired for explanation generation. Handcrafting explanatory knowledge or storing logs of agents' internal states (e.g., mental states and actions) or Prolog of facts appears to be the predominant technique. In this context, resource management for storing explanatory knowledge is a significant concern that is yet to be addressed in the literature. A summary of key findings on deliberative XGDAIs is presented in Table \ref{tab2_supp} in the supplementary materials.

\subsection{Reactive XGDAIs}

From the literature search, models were classified as Reactive for having a set of simple behavioral patterns for reacting to environmental stimuli, without needing to maintain an internal world model of the environment.

\subsubsection{Automated Rationale Generation (ARG) Model - Post-hoc domain-agnostic}

\begin{figure}[ht!]
\centering
\includegraphics[width=0.6\textwidth]{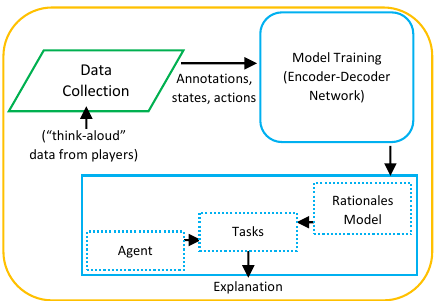}
\captionsetup{justification=raggedright}
\caption{The ARG architecture proposed by \citet{ehsan2019automated} is an end-to-end training method for explanation generation. The technique is to gather a think-aloud data corpus (i.e., actions, game states, and annotations) of players that discussed their actions during the game and train the encoder-decoder network to enable agents to produce reasonable rationales for their actions.}
\label{fig07}
\end{figure}

ARG is an explainable model proposed by \citet{ehsan2019automated} for real-time explanation generation, which is trained to interpret an agent's inner state and data representation in natural language. The approach is domain agnostic and ideally developed for a 'Markovian' sequential environment where the agent's states and actions are accessible to the model and where decisions (i.e., selection of actions that maximize expected future rewards or utility) made by the agent in the past have an impact on future decisions. ARGs are expected to generate explanations or rationales in the natural language for the agent's behavior as if a person had performed such behavior \cite{ehsan2018rationalization}. The idea behind the rationale generation is that people can communicate effectively by verbalizing reasonable motives for their actions, even though the verbalized reasoning is not aligned with the decision-making neural processes of the human brain \cite{block2005two}.

\citet{ehsan2019automated} apply ARG for human-like explanation generation for an agent playing Frogger, a sequential game environment. The ARG approach is to collect a corpus of human-like explanation data (Fig. \ref{fig07}) - i.e., a thought-aloud data corpus of players that discussed their behavior during the game and train a neural rationale generator (an encoder-decoder NN) using the corpus to enable the agents to produce their plausible human-like rationales or explanations.

While the results are promising, the rationale generator's potential limitations may be the lack of a more grounded and technical explanation framework. The framework lacks interactivity that allows users to question a rationale or suggest that an agent clarifies its decision in another way. Another limitation may stem from the data collection process that introduces breakpoints to request an explanation from the players. It is necessary to determine an appropriate method to collect data from players during run-time without interruption of the players or participants.

\subsubsection{eXplainable Reinforcement Learning (XRL) – Post-hoc domain-independent}

XRL is proposed in several studies for a class of reactive RL agents \cite{cruz2019memory}, which are model-free agents that determine what to do depending on their current observations \cite{littman1994memoryless}. They rely on a simple behavioral policy scheme (i.e., a state-to-action mapping) that allows them to learn a policy based on trial-and-error interaction with their environment \cite{sequeira2019interestingness}. RL agents do not generally reason or plan for their future actions, making it challenging to explain their behavior. An RL agent may eventually learn that one action is preferred over others or that choosing an action is associated with a higher value to attain the goal but would lose the rationale behind such a decision at the end of the process \cite{sequeira2019interestingness}. Thus, they lack the mechanism that can effectively explain why they choose certain actions given a specific state. Some existing techniques of explainability for RL agents aim to make the decision process during the policy learning process retrievable and explainable. Some examples include MXRL \cite{cruz2019memory}, MSX via Reward Decomposition \cite{juozapaitis2019explainable}, and RARE \cite{tabrez2019improving}.

MXRL is an XRL strategy proposed by \citet{cruz2019memory} to enable an RL agent to explain to users why it selected an action over other possible actions. The model uses an episodic memory \cite{sequeira2019interestingness} to save each episode or agent's record of executed state-action combinations, then computes both the likelihood of success (Q-values) as well as the number of transitions within each episode to meet the final target to provide an explanation or reason for selecting an action over the others. MXRL is implemented in two discrete simulated environments: a bounded grid world and an unbounded grid environment with aversive regions. The RL agent can clarify its actions to lay users at any time by using information obtained from memory. MXRL suffers from limitations with regard to the utilization of memory in broad solution spaces.

MSX is an XRL strategy proposed by \citet{juozapaitis2019explainable} to explain RL decision agents via reward decomposition. The idea is to break down incentives into amounts of semantically meaningful reward types, allowing actions to be contrasted in trade-offs between them. MSX should provide a concise description of why one behavior is favored over another in terms of reward styles in a domain-independent setting. It uses an off-policy algorithm known as Q-learning that leads to the best policy and decomposed action values. The focus is on explanations that learn Q-functions that allow observing the actions preferences of the agent quantitatively. MSX is implemented in two environments to support its validity: a CliffWorld grid-world where cells can contain cliffs, monsters, gold bars, and treasure that is decomposed into reward types [cliff, gold, monster, treasure] reflecting the current cell's contents; and a Lunar Lander rocket scenario where the actions can be decomposed into natural reward types including crashing penalty, safe landing bonus, main-engine fuel cost, side-engines fuel cost, and shaping reward that defines scenarios close to the actual world of controlling a rocket during a ground landing.

RARE is an extension of the XRL strategy \cite{tabrez2019improving} to address the need for establishing a shared behavior mental model between an RL agent and a human collaborator using a timely update to the reward function. The RARE framework is modeled on the premise that sub-optimal collaboration activity results from a misinformed understanding of the task/assignment rather than a problem with the reasoning of the collaborator. Thus, using the Markov decision-making process, sub-optimal human decision-making is attributable to a malformed policy due to an inaccurate task model. RARE will determine the the most probable reward function for human actions through interactive learning and reward updates. Subsequently, the missing aspects of the reward function is identified and conveyed as actionable information to allow the collaborator to update the reward function (task understanding) and policy (behavior) while performing the task and not after the task is completed. This mechanism should allow the agent or robot to provide a policy update to a person (i.e., by explaining the correct reward to the human) based on the perceived model difference, minimizing the risk of costly or dangerous errors during everyday tasks. A color-based collaborative version of Sudoku and an autonomous robot are used to implement RARE \cite{tabrez2019improving}. The robot is reported to interrupt users who are on the verge of making a mistake, inform them that their actions will cause a failure of the task, and explain which constraint of the game will inevitably be violated. However, the RARE model still lacks the comprehensibility of its optimal policies. During the computation of an optimal policy, the factors taken into account for risk and reward considerations for each state and prospective action are lost.

Other XRL strategies for model-free RL agents include the work of \citet{madumal2019explainable}, which utilizes causal models to generate 'contrastive' explanations (e.g., "why" and "why not" questions) as a means of describing partly measurable agent action or actions in a game scenario (Starcraft II). The approach is to learn a structural causal model (SCM) during RL and to generate explanations for "why" and "why not" questions by counterfactual analysis of the learned SCM. However, one weakness of the approach is that the causal model must be given beforehand. In another work by \citet{pocius2019strategic}, deep RL is used to provide visual interpretations in saliency maps to describe an agent's decisions in a partially observable game scenario. However, saliency maps have several disadvantages \cite{alqaraawi2020evaluating}. They do not help to explain long-term causality. The presence or absence of certain critical features may produce significantly different saliency maps. And finally, saliency maps ignore image features that cannot be localized to individual pixels. The study by \citet{sequeira2019interestingness} provides explanations through an introspective analysis of the RL agent's interaction history. The framework explores the history of an agent's environmental experiences to retrieve interesting elements that explain its behavior.

\subsubsection{Autonomous Policy Explanation (APE) – Post-hoc domain-agnostic}

APE is a strategy proposed by \citet{hayes2017improving} for a class of reactive agent robot controllers that rely on black-box-trained reinforcement learning models \cite{kennedy1998conceptual} or on hard-coded conditional statement-driven policies to enable the robot to autonomously synthesize policy descriptions and respond to natural language queries by human collaborators. The objective is for the robot to reason over and answer questions about its underlying control logic independently of its internal representation, allowing human coworkers to synchronize their perceptions (or mental models) and recognize defective actions in the robot controller. The model applies to discrete, continuous, and complex dynamic environments. Using a Markov Decision Process (MDP) model for constructing the domain and policy models of the control software,the system learns a domain model (i.e., collection of states) of its operating environment and the robot's underlying control logic or policy from actual or simulated demonstrations or observations of the controller's execution traces. These are incorporated into a single graphical model that collects the essential details about states' relationships and behavior.

\begin{figure}[ht!]
\centering
\includegraphics[width=0.9\textwidth]{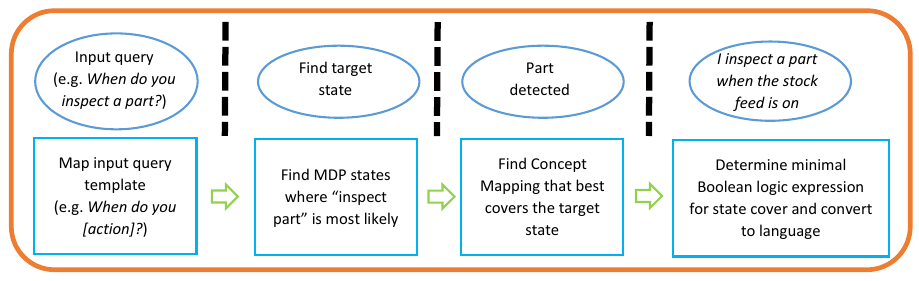}
\captionsetup{justification=raggedright}
\caption{Automated policy generation framework of \cite{hayes2017improving}. The framework maps a policy clarification to an action/input request. An input request or query (e.g. "When do you 'inspect a part'?") is first matched against pre-defined query templates (e.g. "When do you [action]?"). An algorithm for graphical search is used to find the target state, i.e. the state regions that fulfill the query criteria. Using logical combinations of communicable predicates, a concept mapping that best covers the target state is found. The cover is then reduced and translated via the template into language.}
\label{fig10}
\end{figure}

The model employs communicable predicates for natural language communication, i.e., Boolean classifiers similar to traditional STRIPS-style \cite{fikes1971new} planning predicates with associated natural language descriptions, to translate attributes from sets of states to natural language and back (Fig. \ref{fig10}). The algorithms can then enable the agent to answer a direct inquiry about behavior-related questions, e.g., "When do you do \_?" - requesting an explanation about the occurrence of a specific controller behavior, or "Why didn't you do \_?" - requesting an explanation for why a certain controller activity was not observed.

\subsubsection{Summary on Reactive XGDAIs}

Overall, reviewed literature on reactive XGDAIs relies on simple behavioral policy models to generate explanation. Most approaches are agnostic and have the potentials to be applied in several domains. However, several gaps are still visible. For example, some models lack a more grounded and technical explanation framework \cite{ehsan2019automated} (e.g., an interactive explanation platform, breakpoint-free explanatory knowledge collection process). Some others, particularly RL agents, cannot articulate rationales for their actions or 'concerns' (i.e., comprehensibility of agent's optimal policy after policy convergence). A summary of key findings on reactive XGDAIs can be seen in Table \ref{tab3_supp} in the supplementary materials.

\subsection{Hybrid XGDAI}

From the literature search, methods are designated as Hybrid XGDAI if they possess elements of Deliberative and Reactive methods. Hybrid agents typically combine the instinctive responses of Reactive methods with the complex reasoning capabilities of Deliberative methods.

\subsubsection{Perceptual-Cognitive eXplanation (PeCoX) – Domain-agnostic}

\begin{figure}[!b]
\centering
\includegraphics[width=0.5\textwidth]{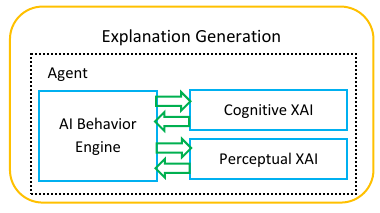}
\captionsetup{justification=raggedright}
\caption{PeCoX generation framework \cite{neerincx2018using}. The framework distinguishes between Perceptual XAI and Cognitive XAI. The Perceptual XAI is connected to the sub-symbolic reasoning of the AI behavior engine and is designed to clarify the perceptual aspect of the agent's behavior. The Cognitive XAI interacts with this part to ground its belief base. It can explain why certain actions have been selected by linking them to goals or beliefs.}
\label{fig11}
\end{figure}

PeCoX is a framework proposed by\citet{neerincx2018using} to create explanations that explore the cognitive and perceptual dimensions of the behavior of an agent (Fig. \ref{fig11}). PeCoX's perceptual XAI explains the perceptual aspect of the agent's behavior using a proposed Intuitive Confidence Measure (ICM) \cite{van2018icm} and a method of contrastive explanations \cite{robeer2018contrastive} that involve the classification of facts and foils, e.g., 'How did you arrive at this outcome (the fact) rather than the other (the foil)?'. PeCoX's perceptual XAI is model-agnostic, relying entirely on any learned model's input, output, and prospective feedback on that output. The ICM (or uncertainty) measures the expected performance of the machine learning model on any specific classification or decision. The perceptual XAI's diagnostic method to probe the agent behavior is thus purely reactive.

PeCoX's cognitive XAI selects goals, beliefs, and 'emotions' to explain why the agent chooses a certain action. This is similar to the BDI explanation models \cite{harbers2010design}. The model considers explanations from the intentional stance \cite{dennett1978three}, i.e., the notion that an agent's action depends on its built-in intention, beliefs, or goals. PeCoX XAI includes another cognitive function: 'emotion', e.g., 'I expect (emotion) you would exercise regularly since I want (goal) you to be physically fit, and I suspect (belief) you are not currently exercising adequately' \cite{harbers2010guidelines} \cite{doring2003explaining}. PeCoX's cognitive framework is designed to be domain-agnostic. The cognitive XAI relies on Ontology Design Patterns for reasoning and communication, and Interaction Design Patterns for shaping multimodal communications. Therefore, the cognitive XAI qualifies as a deliberative XGDAI.

\subsection{Explanation Generation Techniques In Application Scenarios}

To summarize, explanation generation techniques are broadly categorized as Deliberative if the agent intentionally develops a plan based on the current environmental states, Reactive if the agent changes its behavior in response to environmental stimuli, or Hybrid if the agent has elements of both Deliberative and Reactive. This section summarizes some of the explanation generation techniques used in the application scenarios.

Both Deliberative and Reactive explanation generation techniques are used in robot-human collaborative tasks. Deliberative techniques such as the Explainable-Plan technique provide the user with reasoning for the robot's planned methods to which the user can modify the robot's plans as needed \cite{korpan2018toward} \cite{borgo2018towards}. Other Deliberative techniques used in robot-human collaboration include CREATIVE \cite{wicaksono2017towards} for designing tools to meet expectations and Proactive \cite{gervasio2018explanation} to generate explanations in anticipation of unexpected events. APE \cite{hayes2017improving} is one example of a Reactive technique used in robot-human collaboration, while PeCoX \cite{neerincx2018using} is considered a Hybrid technique.

The Explainable NPC \cite{molineaux2018towards} and GDA \cite{molineaux2010goal} are both Deliberative methods in game applications. The former is used to explain agents' actions and motivations within a game context. In contrast, the latter is used to explain unexpected events when observed states do not meet expectations. ARG \cite{ehsan2019automated}, on the other hand, is a Reactive technique for producing human-like rationalization for agents playing a game.

Techniques used in training applications include Explainable BDI \cite{harbers2010design} and Debrief \cite{johnson1994agents}. In both techniques, the mental states and justifications for selecting an action are stored in a log to be replayed if a similar scenario emerges. Explainable BDI produces explanations in the form of the beliefs and goals of the agent making the decision. At the same time, Debrief records the verbal assessments of a human operator making the decisions.

The KAGR technique used in MAS \cite{sbai2013intra} formulates agent reasoning for specific events as a quadruplet of Knowledge, Action, Goal, and Relation.

\section{Explanation Communication Techniques for XGDAI}

This section discusses what type of explanations (e.g., logs, text) are communicated from the agents to users and how they are communicated (e.g., visual, textual, speech). This section distinguishes explanation communication for XGDAI into visual, verbal, gestural, numerical, and textual expressions. Our findings are summarized in Table \ref{tab4_supp} in the supplementary materials.

\subsection{Visualization}

Visualization is a technique of communicating an agent's plan, "mind", or decision-making system by externalizing the pathways of the decision support process. This technique builds visual mediums between the agent and the humans to establish trust and transparency between humans and machines. An important justification for plan visualization, for instance, is the need to minimize the time taken to communicate the agents' plans in natural language to the humans in the loop. In some domains, visual explanations are well suited to be integrated into the typical workflow of experts: for example, in the medical domain, experts are used to analyze the results of imaging procedures (X-ray, fMRI, etc.), drawing their attention to potentially relevant parts of the image can be an intuitive interface.

Visualization techniques in this section are further divided into NN approaches and symbolic approaches. NN approaches provide low-level explainability of a model's internal representation using visual cues and, otherwise, leave interpretation up to the user. In symbolic approaches, knowledge is encapsulated into a user-interpretable format, i.e., symbols directly computed using traditional connectionist models.

\subsubsection{Neural Network Approaches}
\subsubsection*{Class Activation Map (CAM)}

Deep learning architectures based on convolutional neural networks (CNN) achieved optimal results in many computer vision applications. From the training data, CNNs learn to extract a deep hierarchy of task-relevant visual features \cite{townsend2019extracting, das2020opportunities, arik2020explaining, kori2020abstracting}. While these feature-extracting filters can be visualized, they are hard to interpret: in lower layers, the filters are mostly edge detectors, while in higher layers, they are sensitive to complex features. Moreover, the filters represent what image features the CNN is sensitive to, but not what features lead to a given image classification. CAMs \cite{zhou2016learning} address this issue by creating a heatmap of discriminative image regions over the input image that shows what parts of the input image contributed how much to the CNN's classification of the image as belonging to a selected class. In this way, CAMs supply a visual explanation for a classification. Furthermore, by calculating and comparing CAMs for different classes, relevant portions of the input image can be visualized for distinction.

CAMs work by inserting a global average pooling (GAP) layer directly after the architecture's final convolutional layer. This layer computes a spatial average of all filters from the previous layer, which are then weighted by the selected output class. A drawback of this approach is that the NN architecture is altered, and models need to be retrained. Gradient-weighted Class Activation Mappings (Grad-CAM) were introduced by \citet{selvaraju2017grad} based on gradients flowing into the final convolutional layer to solve this problem. They extend their approach further by fusing it with Guided Backpropagation (Guided Grad-CAM) to enhance the resolution. An example application field for CAM and related methods is medical image analysis. \citet{ng2018classification} use a three-dimensional CNN to analyze MRI data of possible migraine patients. They use CAMs to highlight discriminative brain areas. In such applications, CAMs guide a medical expert's attention to different image regions to assess the network's prediction. 

\citet{kerzel2022whats} used Grad-CAM as a means for explaining neural robot grasping. Given an object to be grasped, a motor control network generates a movement command for a robot arm to pick up the object. Grad-CAM is then used to highlight the parts of the object which are relevant for generating the movement for a particular joint motor.

\subsubsection*{Attention Weights}

Recurrent neural networks (RNNs) are a popular deep learning framework for temporal modeling, particularly in medical applications for modeling disease progression \cite{esteban2016predicting} and predicting diagnoses based on patient records \cite{lipton2015learning}. One solution to improve interpretability or visualization was introducing an intermediate attention layer \cite{choi2016retain} \cite{sha2017interpretable}. The attention mechanism decomposes a complex input into a series of accumulated attention weights. For example, the Timeline model \cite{bai2018interpretable} aggregates contextual information of patient medical codes using an attention mechanism to predict future diagnoses of future medical visits. Medical practitioners were able to gain insight and anticipate future diagnoses by observing the attentional weights of medical codes over time. Theoretical scenarios can be simulated by manipulating attention weights and observing the subsequent prediction changes using attention-weight visualization tools \cite{lee2017interactive} \cite{liu2018visual}.

For instance, neural machine translation (NMT) uses tree search to translate a sentence from one language to another \cite{lee2017interactive}. A search decoder determines which candidate words are the most appropriate translation based on the attention weights for specific words. A candidate-translated sentence is therefore generated by selecting specific candidate words that maximize the overall attention weights. The authors provided a visualization tool where the user can manually select other candidate words, to which the system recomputes the attention weights for subsequent candidate words to complete the sentence. Color intensity is used for highlighting words with the highest attention, making it easier for users to observe the entire candidate sentence.

\subsubsection*{SHapley Additive exPlanation (SHAP)}

\begin{figure}[!tb]
    \centering
    \includegraphics[width=0.65\linewidth]{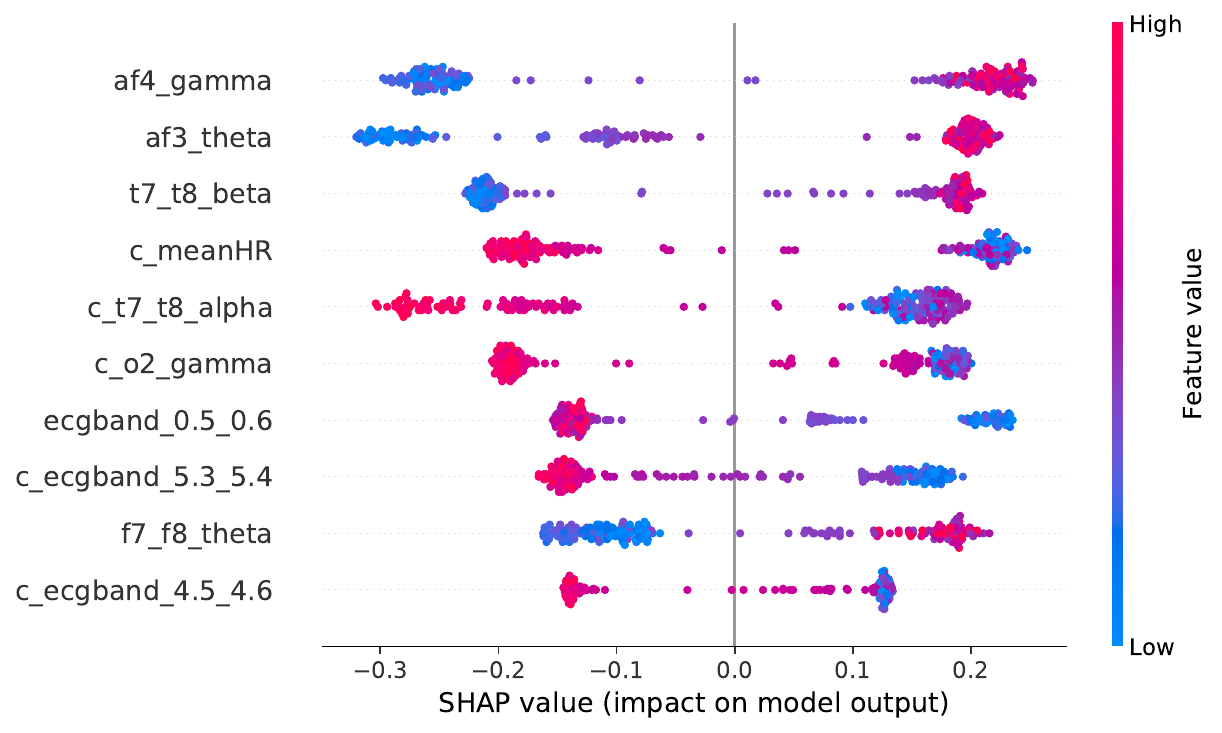}
    \includegraphics[width=0.65\linewidth]{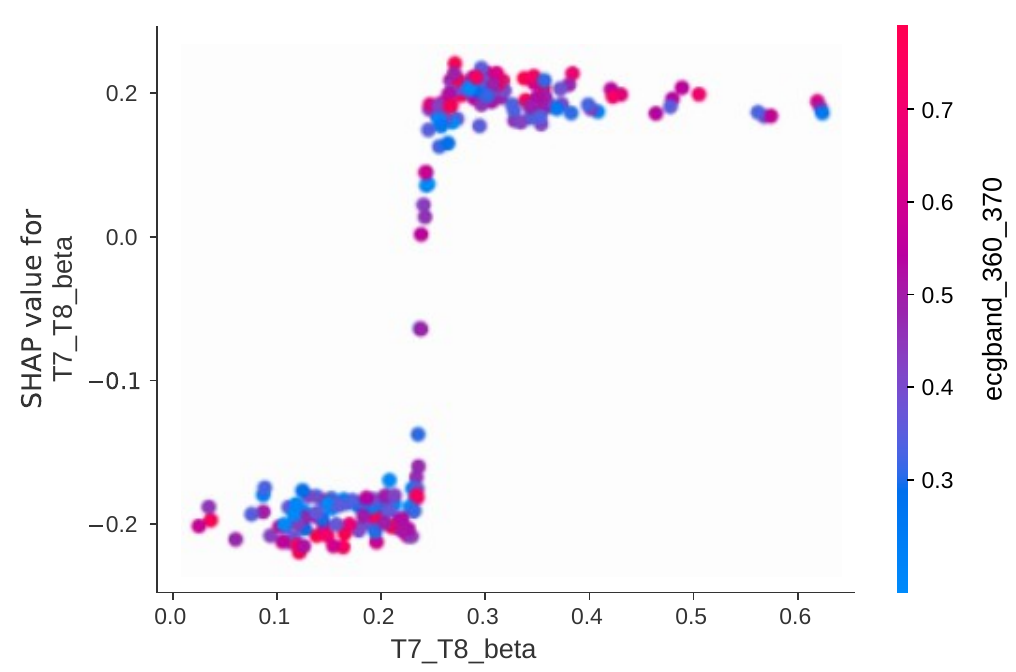}
    \caption{Example of a SHAP summary plot (top) and feature dependence plot (bottom) describing feature importance for affect recognition \cite{liew2021emotion}.}
    \label{fig_shap}
\end{figure}

\citet{lundberg2017unified} utilize SHapley Additive exPlanation (SHAP) values, an extension of Shapley values, as a united strategy to explain any machine learning model output. Given a collection of data points for training a prediction model, SHAP values are computed to determine the impact of a data point or feature on the model's predictions, giving the cumulative marginal contributions of individual inputs and features. SHAP values can thus be considered as a set of "importance scores". As the SHAP values are discretely computed for inputs and features, approximations of global input importance, or the global importance of a feature, can be easily calculated. Feature importance can be evaluated globally by taking the combined SHAP scores for a particular attribute or interaction with other features using a dependence scatter plot. The discrete SHAP values can be easily visualized, as shown with examples in Fig. \ref{fig_shap}, providing users with easily interpretable information. The top sub-figure highlights the relation of the feature values (i.e., the color of the dots) and the impact of their contributions towards the model output based on their positions on the horizontal axis. The bottom sub-figure represents an example of an interaction plot between two features with each dot representing one data point. The combination of dot colors and their positions on the x-y axes forms an illustrative representation of the relation between the two features. 

\subsubsection{Neural-symbolic Approaches}

Representations in NN are distributed among many neurons, making them difficult to isolate and interpret without relying on post-hoc knowledge extraction methods. Symbolic models such as graphs and decision trees are easier to interpret, as each node or symbol represents a single distinct concept. Neural-symbolic (NS) models combine the efficient learning and inference power of traditional NN and the interpretability of symbolic knowledge extraction and reasoning \cite{townsend2019extracting} \cite{garcez2019neural}. 

\subsubsection*{Visual Grounding via NS-structured Graphs}

Visual grounding is the general task of locating a structured description's components in an image \cite{conser2019revisiting}. The structured description or query can be a natural-language phrase that has been parsed as a symbolic structured graph such as a scene graph or action graph. In human-robot interaction (HRI) scenarios, visual grounding can enable a robot to accurately locate an object amongst other objects in the workspace \cite{shridhar2020ingress}. Scene graphs were first used for the visual-grounding challenge by \citet{johnson2015image}. The scene graph captures the visual scene's intricate semantics by expressly modeling objects, their descriptive properties, and their interconnections in the scene. For example, a scene described by the sentence "man riding black horse" is converted to a scene-graph representation with nodes corresponding to objects ("man"), descriptions or attributes ("horse is black"), and interconnections or relationships ("man riding horse"). The grounding task creates bounding boxes corresponding to the specified objects, such that the located objects have the specified attributes and relationships.

An NS "end-to-end" approach to visual grounding was proposed by \citet{xu2017scene} to generate structured scene representations from input images. The model addresses scene graph inference by using conventional RNNs that learn to generate image-grounded scene graphs and iteratively improve their predictions via message passing. A Region Proposal Network (RPN) \cite{ren2016faster} is used to generate a collection of object proposals from an image and transfers to the graph-inference module the extracted features of object regions. It then outputs a scene graph consisting of categories of objects, semantic relations between object pairs, and their bounding boxes. A significant aspect of the work is the message passing, of contextual information between two sub-scene graphs (bipartite graphs) and how RNNs can refine their predictions. 

In the same context, \citet{yang2018graph} present Graph R-CNN, a model for constructing scene graphs that can also perform object detection and infer object relations in images. The main framework consists of a Relation Proposal Network (RePN) dealing with the quadratic number of possible relationships between the objects and an attentional Graph Convolutional Network (aGCN), which extracts context information between objects and their relations. 

\citet{zhou2018weakly} propose a weakly-supervised framework for reference-aware visual grounding for instructional videos. The only thing supervised is the temporal synchronization between the video segment and the transcription. The framework consists of a visually grounded action graph, i.e., an optimized reference-aware multiple instance learning (RA-MIL) objective for poor control of grounding in videos. A standardized representation captures the latent dependence between grounding variables and references in videos. Scene graph or action graph representations hold tremendous promise in visual grounding tasks, particularly for enhancing an agent's perceptual-cognitive framework. However, extracting scene graphs from images is still a significant challenge.

\subsubsection*{Extraction into Automata and Transducers}

Finite-state machines or automata (FSAs) are computational models that assume one of any number of finite states at any time and transition to other states in response to particular inputs. Architectures of FSAs are inherently interpretable and illustrated by a series of states linked by transition functions. A large body of work exists for extracting interpretable FSAs from RNNs. For example, using quantization algorithms \cite{giles1992learning}, clustering outputs to infer FSAs \cite{zeng1993learning} \cite{omlin1996extraction}, and recursively testing inputs and outputs \cite{weiss2018extracting}.

Transducers are FSAs with both an input and an output, whereas traditional FSAs have only an input. Transducers can be further divided into Moore machines and Mealy machines. The former produces outputs based solely on the machine state, while the latter relies on both inputs and states. Preference Moore machines \cite{wermter1999preference} can operate as either a simple RNN or as a symbolic transducer while integrating different neural and symbolic knowledge sources. In a text-mining application \cite{arevian2003symbolic}, symbolic transducers were able to encode concepts from a news corpus quickly. At the same time, neural machines were able to produce improved classification performance with additional training. Transducer extraction \cite{wermter2000knowledge} was then able to obtain detailed symbolic information while preserving sequential information.

\subsubsection*{Rule Extraction Algorithms}

Rule extraction algorithms are a method for converting a neural network's internal arrangement of neurons into a linguistically interpretable format. The simplest form of a rule can be described using a conditional "IF-THEN" statement, represented in a NN by a single neuron as "IF input=X, THEN output=Y." \citet{andrews1995survey} classify rule extraction algorithms into three categories. Rule extraction algorithms that work at the neuron-level instead of the whole NN level or global are called decompositional methods. If the NN is a black-box model (i.e., activation or weights of the network cannot be observed directly), then the algorithms are pedagogical. The third category, eclectics, is a combination of decompositional and pedagogical algorithms.

The Local Rule EXtraction (LREX) method \cite{mcgarry2001knowledge} is an example of eclectical rule extraction from radial basis function (RBF) networks. LREX examines the hidden unit to class assignments and extracts rules based on the input space to output space mappings. An example of decompositional rule extraction is the technique proposed by \citet{malone2006data} extracts propositional "IF-THEN" type rules from self-organized mapping networks.

Another class of rule extraction bypasses the need for neural networks and directly converts entire data sets into a more condensed set of rules. \citet{mcgarry2007auto} developed a method for automatically generating biological ontology and integrating them into Bayesian network frameworks to enable reasoning and prediction. \citet{mao2019bootstrapping} introduced an algorithmic methodology for encoding relational representations of visual inputs in the form of sparse knowledge graphs. \citet{sutherland2020tell} introduce a context-agnostic methodology for extracting target-sentiment-cause triplets from sentences.

Given many data samples, extracting interpretable rules is a balancing act. Simplifying rules by discarding outliers may produce an oversimplified rule set that may not respond correctly to rare events. \citet{sovrano2021explanation} propose a concept called "eXplanation Awareness" (XA) for organizing acquired experiences into clusters labeled on a case-by-case basis. XA enables sampling experiences in a curricular and task-oriented manner, focusing on events' rarity, importance, and meaning.

\subsection{Textual And Verbal Communication}
\subsubsection*{Textual Data Logs}

For agents that rely on relational (or symbolic) logics for knowledge representation and reasoning (KRR), using logs of data comes easily to communicate the internal state of the robots. An example is the explainable BDI model proposed by \citet{harbers2010design}, where all past mental states and actions of the agent needed for explanations were stored in a behavior log. When requesting an explanation, the algorithm selects beliefs and goals from the logs for the explanation. Other examples are the XAI architecture proposed by \citet{van2004explainable} for NPCs in Full Spectrum Command, which extracts main events and decisions from the replay log during the post-review process.

\subsubsection*{Programming in Logic (Prolog)}

Prolog is a general-purpose logic programming language. An important aspect of this language is the utilization of symbolic representations - e.g., "?-dog(jane). no" corresponding to "Is Jane a dog? No - a cat". The Prolog program uses rules and facts and queries that search through stored facts and rules to work out answers \cite{bramer2005logic}. \citet{wicaksono2017towards} use Prolog for the explainable CREATIVE model to express and store relevant information needed for explanations, such as acquired hypotheses, primitive object poses, and camera snapshots. CREATIVE represents a set of simple questions and answers and describes the relationship between objects as Prolog facts to enable explanation.

\subsubsection*{Verbal Communication}

Speech or verbalization has been one of the earliest means of communicating agents' thoughts, beliefs, or actions, especially in the domain of social or service robots. Since social robots must converse daily with humans, they require skills for a natural conversation ability. Speech offers a more intuitive and faster means of explanation communication (compared to text) for robots, particularly interacting with the visually impaired or children who have not yet developed a full reading competency. The earliest generation of humanoid robots (i.e., developed in the 1970s and 1980s) was equipped with conversational skills, although the skills were usually primitive. They were typically designed as simplistic input/output mapping speech combinations \cite{breazeal2008social}. In a study conducted by \citet{ambsdorf2022explain}, participants' perception towards robots were evaluated as they watched two NICO robots played tic-tac-toe against each other. While both robots utilized verbal communication, one robot only announced its moves while the other robot explained its reasonings for choosing a move. The explaining robot was perceived to be more human-like and lively compared to the non-explaining robot, which suggested that verbal explanations are a viable method to foster trust in human-robot cooperation tasks. 

Another good example is the Waseda Robot, WABOT-1 \cite{shirai1973ear}, which was designed with the capability to recognize spoken sentences as concatenated words, make vocal responses, and change a related state using a Speech Input-Output System (SPIO) \cite{shirai1974algorithm}. The WABOT-1 system could accept Japanese spoken command sentences only in sequences of individually spoken Japanese words and then respond to the meaning of the command verbally to make the robot move as commanded. The system's inner core works as an automaton, making output and transitions after recognizing an input sentence. A further upgrade on the conversational system to make the speech more natural, particularly the speech synthesis part, was introduced in WABOT-2 \cite{kobayashi1985speech}, a robot musician which could produce speech responses by retrieving a word dictionary corresponding to a code of the spoken command. The dictionary stores the names of consonant and vowel syllabic units necessary for each unit's words, vowel durations, and accent patterns.

In this class, another more recent social robot is PaPeRo \cite{osada2006scenario}, a childcare robot designed to converse with humans more naturally. PaPeRo uses a dictionary of commonly used words and phrases and could also be updated by the designers. Humans interacting with the robot must also converse in similar words and phrases that the robot understands to enable natural conversation. PaPeRo uses an electronic hardware auditory system for human-robot communication \cite{sato2006auditory} and can recognize multiple utterances, give a quiz to children who provide answers to the quiz using a special microphone, and tell in natural language the names of the children who got the correct answer. Other similar robots are Honda Asimo \cite{sakagami2002intelligent} which uses a commercial hardware electronic system for speech synthesis, and ASKA \cite{nisimura2002aska} receptionist robots, with a conversational speech dialogue system that can recognize users' questions and answer the users' questions by a text-to-speech voice processing with other additional intuitive channels such as a hand gesture, head movement, and body posture. Robovie \cite{sabelli2011conversational} is another example in this category that can communicate in English using a vocabulary of about 300 sentences for speaking and 50 words for recognition. The reader is referred to \citet{leite2013social} for a more comprehensive review of social robots.

\subsubsection*{Referring Expression}

Referring expression is a language construction or expression used to identify particular objects within a scene \cite{yu2017a}. Generation of Referring Expressions (GRE) or Referring Expression Generation (REG) is expression selection for identifying an element from a set of possible elements when given a shared context between a speaker and the hearer \cite{foster2008the}. REG algorithms have traditionally been applied as computational models of people's ability to refer to objects \cite{deemter2012generation}. Many scholarly works on REG can be found spanning the last five decades \cite{krahmer2012computational}. In recent years, REG has received significant attention in HRI scenarios, particularly for situated human-robot dialogues. In situated dialogue, humans and robots/agents are co-present in a shared physical environment or workspace (e.g., in assembly tasks) and often need to refer to an object in the environment that is being shared or manipulated \cite{fang2013towards}. The motivation for REG is thus the need for the robot to comprehend expressions concerning objects in a shared space and their relationships from inputs of image and natural language. In a contribution by \citet{giuliani2010situated}, a situated reference generation framework is presented for an HRI system to collaborate with humans in an assembling task that involves building objects from a set of wooden toys. The REG system includes a goal inference sub-symbolic system that can identify people's goals or mistakes by observing their verbal and non-verbal actions. The robot's REG module then uses situated references to clarify to users the mistakes and suggest strategies for solutions \cite{dale1995computational}.

Another relevant example is the INteractive visual Grounding of Referring ExpreSSions (INGRESS) framework proposed by \citet{shridhar2020ingress}. INGRESS is a robot that executes instructions to pick and place items (in the environment) in natural language. The system can ground referring expressions, i.e., recognizing dialogues or references to objects and their relationship from images and natural language inputs. It may also ask questions for interactive clarification of unclear referential expressions. It uses a two-stage long short-term memory (LSTM) framework to establish visual representations of objects and link them to input language expressions to identify the candidate referred-to-objects.

Some other relevant contributions are the gestural-deictic reference by \citet{kranstedt2005incremental} and the haptic-ostensive reference by \citet{foster2008the}, in which the authors argue that a multimodal reference involving manipulation actions with the objects would provide a richer referring approach for conversational partners in situated dialogue. They suggested that a fully elaborated linguistic reference may not always be necessary between conversational partners.

\subsection{Explanation Communication Techniques In Application Scenarios}

In summary, explanation communication techniques are grouped by explanatory forms: visual, numerical, verbal, and textual. This section summarizes some of the explanation communication techniques used in application scenarios. Communication techniques used in robot-human collaborative tasks mainly include human-like expressive ability such as gestures \cite{mikawa2018expression}, motion \cite{mikawa2018expression}, and verbal communication \cite{kobayashi1985speech, osada2006scenario, nisimura2002aska, sabelli2011conversational, yu2017a}. The familiarity of human-like communication is more intuitive and user-friendly for people new to interacting with intelligent robots. Similarly, simple communication methods such as expressive lights \cite{baraka2016expressive} \cite{song2018effect} are unexplored territory for XGDAI. 

One domain where AIs are highly in demand is e-health for applications such as AI-assisted medical diagnosis caretaker robots. Visualization methods such as CAM \cite{ng2018classification} highlight regions of interest that are relevant to the diagnosis. Numerical methods such as attention weights \cite{lipton2015learning, choi2016retain, sha2017interpretable} are used for fine-tuning diagnosis algorithms to minimize misdiagnoses. In scenarios where AIs are used for training scenarios, XGDAI helps to diagnose the how and why the agent makes a particular decision during field exercises. Logged recordings of the agents' actions and their rationales \cite{harbers2010design} \cite{van2004explainable} can then be parsed to identify and correct any behavior that may produce undesirable outcomes.

\section{Continual Learning for Explainability}

In this section, we examine techniques in XGDAI that enable continual learning of explanatory knowledge, domain knowledge, domain models, or policies (e.g., sets of environment states) for explanation generation. This section explores the solution to two issues: handcrafting domain knowledge artifacts for explanation generation in deliberative symbolic agents; and learned policy losses during the decision-making process for explainability in reactive RL agents.

\subsection{Case-Based Reasoning (CBR)}

For agents that rely on handcrafted domain knowledge for defining the explanation components (e.g., expectations, discrepancy in definitions, knowledge of how to resolve the discrepancy, operator feedback), the common challenge is that substantial domain engineering is done on the system which needs to be updated each time the robot changes to a new environment. CBR is used in many XGDAI approaches to acquire domain knowledge at run-time to minimize the prerequisite domain engineering for generating explanations. CBR applies a learning and adaptation strategy to help explanation generation by storing, recalling, and adapting knowledge information or experiences (or cases) stored in a case-library or memory \cite{urdiales2006purely}.

In the CBR framework, agent learning to expand its knowledge is executed by evaluating and integrating new experiences into the case-library and/or re-indexing and reusing previous experiences \cite{kolodner2014case}. CBR has two primary learning methods: observational learning (supervised) and learning through personal experience \cite{urdiales2006purely}. Observational learning is performed by filling the case-library with observations from expert demonstrations or actual data \cite{van1998learning}. Learning from one's personal experience \cite{kohonen1990self} happens after a reasoning process that evaluates a potential solution to a challenge. If the solution succeeds, it is saved and applied for future references \cite{urdiales2006purely}.

\citet{weber2012learning} propose an observational learning approach to reduce the number of domain knowledge acquisitions necessary to implement the real-time strategy game StarCraft. CBR was applied to learn expectations, explanations, and goals from expert demonstrations. An adversary library offers examples of adversary behavior, and a goal library chooses goals to be pursued by the agent. The learning system offers explanations to the agent if it achieves an intended goal or if a deviation is observed. In a similar effort, \citet{floyd2016incorporating} applied two CBR systems to assist agent learning and explanation generation. Both CBRs use cases learned while interacting with the operator, i.e., learning by observation. The first CBR process determines if a robot is trustworthy and selects a new behavior if found otherwise. The second CBR is used to provide an explanation or clarification should the robot change its behavior. The agent's explanations are based on explicit feedback received from an operator. The model is evaluated in a simulation environment involving an operator who instructs the robot to patrol, detect suspicious objects, and designate the objects as either harmless or dangerous.

\subsection{Explainable Reinforcement Learning}

One of the major concerns for many RL agents is a loss of information about the decision process during the agent's policy learning process. An RL agent may know at the end of a learning objective that certain actions may produce a higher gain or value to attain the goal or that one action is preferred over another but loses the rationale behind the decision-making as the policy converges towards an optimal mechanism for the selection of actions \cite{sequeira2019interestingness}. Once an optimal policy has been learned, the lack of bookkeeping, traceability, or recovery of this process makes it hard for the agent to explain itself or transfer a learned policy for explainability. A few extensions to applying the XRL technique seek to enhance retention of the learned policy for explaining agents' decision processes. Some examples include MXRL proposed by \citet{cruz2019memory} which introduces an episodic memory to store important events during the robot's decision-making process. The episodic memory enables the agent to introspect, observe, or analyze its environment transitions and interactions. The agent is shown to clarify its actions to lay users during task execution, relying on its episodic memory. The major shortcomings in the MXRL technique, as with many other memory-based continual learning techniques, are the limitations with regard to the use of memory in large solution spaces. There is still an open-ended quest for XRL techniques that enable comprehensibility, continual learning, and policy retention.

\section{Discussion}

\begin{figure}[!t]
\centering
\includegraphics[width=0.5\textwidth]{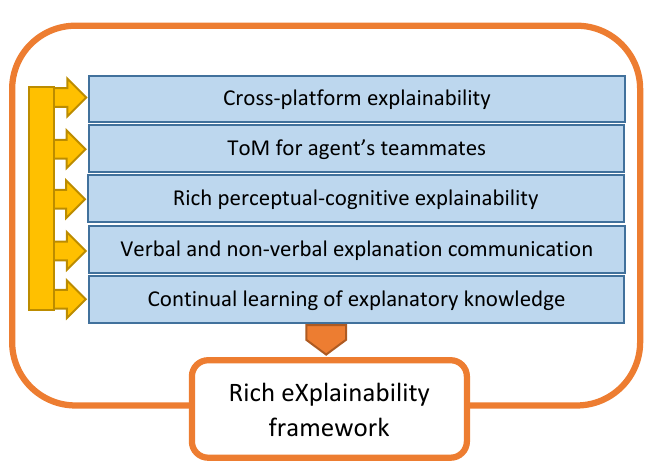}
\caption{Explainability Framework for XGDAI.}
\label{fig17}
\end{figure}

Existing studies in XGDAI show a lack of consensus on the requirements for explainability. Different behavioral architectures for GDAI - e.g., deliberative, reactive, and hybrid - come with different explanation generation techniques.The state of the art suggests the need for an effective unified approach towards explainability in XGDAI. Many explainability techniques still lack an extensive framework consisting of a rich perceptual-cognitive explainable framework, verbal and non-verbal communication framework, framework for natural language processing, and continual learning for explanation construction. In this section, we outline a framework, illustrated in Fig. \ref{fig17}, for the effective actualization of explainability in XGDAI and a road map in Fig. \ref{fig_roadmap} illustrating the four phases of explanation in future research.

\subsection{Framework For Explainability In XGDAI}
 
\subsubsection{Cross-platform Explainability}
 
In the current state of the art, explanation techniques, particularly for symbolic deliberative XGDAI, are domain-specific, relying heavily on the domain knowledge (or learning of the domain knowledge) for constructing explanations. In many such applications, agents already have plans and a clear decision system to achieve a goal; thus, constructing a rich explanation is often relatively straightforward. However, applying these techniques can be problematic as agents perform optimally in a specific domain and sub-optimal in other domains where knowledge of the agent's world is not fully represented in the framework. A cross-platform explainability framework can significantly benefit existing work to improve agents' performance and minimize re-engineering of the domain knowledge.
 
Some emerging techniques, such as the domain-agnostics approaches, suggest a useful notion of explainability that enables cross-platform explanation generation for agents and robots. Most of these techniques can be found for reactive black box agents whose decisions are based solely on the current environment state, not defined \textit{a priori}. Without a world or domain knowledge model, these agents can generate explanations based on the policy learned. However, these platforms are less extensive, and a significant research effort is still required. A cross-platform explainability approach should significantly benefit both deliberative and reactive agents. Also beneficial would be approaches for bridging the gap between symbolic and reflexive (neural network) approaches to explanation generation and communication.

\subsubsection{Theory of Mind (ToM) for Agents' Teammates}

ToM tends to reason about other peoples' perceptions, beliefs, and goals and take them into account \cite{byom2013theory}. A significant body of work on XGDAI involves agents/robots collaborating with humans and other agents. Given this reality, it is imperative for agents to adequately understand their teammates for effective collaboration and provide a useful and timely explanation when necessary. A useful step in this direction may be to integrate the ToM concept in the explanation framework enabling an agent to also reason about other teammates' perceptions and mental states. A well-constructed ToM should enable the robot to understand its teammates' expectations and provide a useful, relevant, and timely explanation. The motivation here is that humans are well known to collaborate with their teammates extensively and explain their behavior using the ToM concept. Currently, only one approach mentioned generating explanations about user/team-member expectations and possible violations of expectations \cite{gervasio2018explanation}.

\subsubsection{Rich Perceptual-cognitive Explainability Framework}

A significant number of approaches, particularly those on deliberative XGDAIs, explain at the level of agents' cognitive functions, which are not grounded on actual agents' perceptions of the real world. On the other hand, a few studies, mainly reactive XGDAIs, highlight procedures for explanation generation at the level of the agent's perceptual function (sensor information, environment states, etc.) with a poor or non-existent explainable cognitive framework or explainable decision-making framework. A rich perceptual-cognitive explainable framework that abstracts low-level agent perception (primarily perceptual explanatory knowledge) for high-level cognition and explainability would significantly advance the current research on XGDAI. Some emerging techniques of explainability/interpretability would be useful, e.g., CAMs to identify the salient information in an image, visual grounding to localize an image component, and REG to enable an agent to distinguish between objects in a shared workspace (e.g., in situated human-robot dialogues).

\subsubsection{Natural Language Processing (NLP)}

NLP is the computer system or AI system's ability to understand, analyze, manipulate, and generate human language \cite{indurkhya2010handbook}. NLP is crucial in the explanation generation/communication framework for XGDAI to enable a human-comprehensible explanation of an agent's/robot's perception, cognition, or decisions. Currently, the state of the art in XGDAI reveals less extensive or even a non-existent NLP ability for many of the agents and robots surveyed. Adding a rich NLP system to existing frameworks would significantly benefit XGDAI in its usefulness and applicability.

\subsubsection{Integrating Verbal and Non-verbal Explanation Communication}

The current state of the art reveals different explanation communication modalities for XGDAI applied in separate niche areas/scenarios. With diverse application scenarios, the need to communicate the agent's/robot's plan, decision, intentions, etc., by combining both verbal and non-verbal communication means is necessary if such an explanation should be natural and effective. A relevant example is seen in how humans explain/communicate using verbal and non-verbal communication, depending on which is most effective for the circumstance. XGDAI should also enable such a combination of different modalities. 

For example, for agents sharing a pedestrian walkway with humans, rotational head/eye movements seem more natural and effective to communicate the agent's decision to turn left, change paths, etc. \cite{mikawa2018expression}. Simultaneously, verbal communication would also prove useful to alert other pedestrians of the agent approaching when they are not aware of its presence. For collaboration with teammates, speech for normal explanatory conversation is necessary. However, expressive motion like gesture (e.g., head nodding) would be useful to provide an implicit explanation, or an expressive light \cite{baraka2016expressive} \cite{song2018effect} or sound to alert teammates of its mental state in an emergency. \citet{kerzel2022whats} proposed a framework which integrates multiple types of explainability methods in a neuro-robotic architecture for object recognition and grasping: first, an LED-based facial expression on the robot’s face is used to give a human interaction partner quick and intuitive feedback about understanding the given task (positive emotion expression) and possible issues (negative emotion expression); second, a more detailed verbal explanation of the object recognition outcomes is generated; and third a visual-based Grad-CAM is used show the user how different areas of the scene, as perceived by the robot, correlate with specific joint motor movements while grasping the object. There can be many possible effective and natural combinations of explanation communication modalities. More research work is still required to bridge this gap.

\subsubsection{Continual Learning of Explanatory Knowledge}

As agents/robots interact with their environment, teammates, and supervisors, they are expected to explain their decisions or actions in different situations and scenarios. Handcrafting explanatory knowledge to satisfy all possible situations and expectations is difficult and requires significant effort or domain engineering. In this respect, an agent's ability to continuously learn to generate/construct explanations in different situations is crucial to explainable agents' success. As in traditional machine learning, learning could be achieved by supervised learning (e.g., learning from demonstration \cite{van1998learning}), unsupervised learning \cite{kohonen1990self}, and reinforcement learning. CBR \cite{urdiales2006purely}, and XRL \cite{cruz2019memory} techniques have been applied in a few studies on deliberative and reactive agents, respectively, to address this concern. However, for reactive XRL, a major concern is how to retain policies learned by agents for constructing explanations. There is still a significant gap to fill in this direction. Scalability and resource management issues would also need to be addressed if explanatory knowledge is stored in a continual learning framework.

\subsection{Road Map}

\begin{figure}[t!]
\centering
\includegraphics[width=0.95\textwidth]{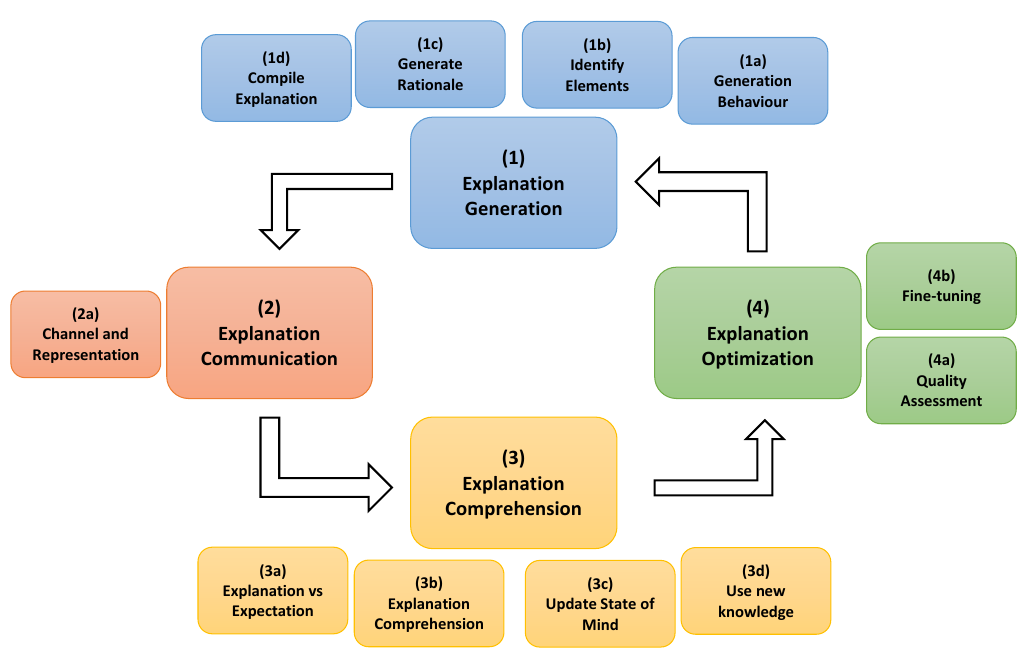}
\caption{Four phases of explanation. The explanation cycle begins by generating an explanation (1). The algorithms' behavior determines when an explanation should be generated (1a), followed by identifying the elements needed to construct the explanation (1b). The explanation and the rationale are then compiled (1c-1d). The channel and representation determine how the explanation is conveyed (2a). The recipient compares the received explanation against its expectations (3a) and attempts to fit the new information into its world model (3b-3c). Subsequent actions then take into account the new information (3d). The performance of the agent with the updated information is assessed compared to its previous performance (4a). Using optimization algorithms such as genetic algorithms, the effectiveness of the agent is incrementally improved (4b) by adjusting the explanation parameters in (1), (2), and (3).
}
\label{fig_roadmap}
\end{figure}

This section proposes a road map for extending research into XGDAI, based on the representation shown in Fig. \ref{fig_roadmap}. The explanation cycle is categorized as the four main phases Explanation Generation, Explanation Communication, Explanation Comprehension, and Explanation Optimization. 

\subsubsection{Explanation Generation}
This phase concerns the elements needed to activate and generate explanations. The explanation generation phase is dependent on the behavior of the AI of the agent or robot. The key research directions of this phase are as follows: 
\begin{itemize}
    \item [(1a)] Existing works focused on advancing the capabilities of the agent (i.e. cognitive and behavioral architecture). However, most methods do not support explainability functions. To further research into XGDAI, equipping the agent's internal AI architecture with a mechanism to generate explanations is needed.
    \item [(1b)-(1d)] Although \citet{kaptein2017personalised} describe how context-awareness and personalization are essential for XGDAI, few works in the literature fulfil the criteria. There is a need for new techniques to facilitate (1b) identification of elements relevant for an explanation; (1c) determining the explanation rationale; and (1d) assembling the elements into a sound explanation.
\end{itemize}

\subsubsection{Explanation Communication}
This phase is needed to convey the generated explanation in an understandable format to the other party, which can be another agent or a user. The identified research directions are:
\begin{itemize}
    \item [(2a)] Explainable agents are likely to be deployed in diverse environments. Therefore, there is a need for the agent to support more than one type of communication channels (i.e. visual, auditory, expressive). Approaches in this direction are rarely found in the literature even though this is an essential element to effectively and efficiently convey explainable information. Although depending on the application scenario, comprehensive explanations can be conveyed using simple communication channels such as expressive lighting signals \cite{baraka2016expressive} \cite{song2018effect}.
\end{itemize}

\subsubsection{Explanation Comprehension}
In this phase, the receiver receives the generated explanation through the chosen communication channel to be converted into useful information. The following points should be considered:
\begin{itemize}
    \item [(3a)-(3b)] Metrics should measure the efficiency of the generated explanation as well as the usefulness of the explanation to the receiver. 
    \item [(3c)-(3d)] The agent/robot keeps track of a model of the user knowledge. The model is regularly updated to reflect the growth of the user expertise and how the user views the State of Mind (SoM) of the agent/robot. 
\end{itemize}

\subsubsection{Explanation Optimization}
Completing the explanation cycle, this phase tracks the explanation flow from beginning to end to identify elements that can be fine-tuned to improve the quality of future explanations. To that end, the following research directions are identified:
\begin{itemize}
    \item [(4a)] Several elements in the explanation cycle can be changed to improve the effectiveness of the explanations. For example, depending on the user feedback in (3), explanation elements in (1) can be altered to exclude redundant information. In addition, the user can query for more information leading to a different choice of explanation generation technique in (1d) or choice of communication medium in (2). The modifications are then evaluated to determine if the quality of the explanations has improved.
    \item [(4b)] Based on the compiled information in (4a), an optimization algorithm such as evolutionary algorithms or reinforcement learning is used to modify elements in the explanation cycle to improve the effectiveness of future explanations.
\end{itemize}

\begin{figure}[!t]
\centering
\includegraphics[width=0.9\textwidth]{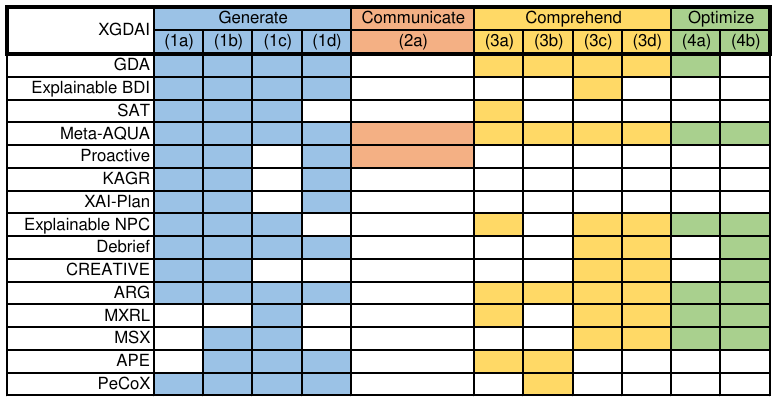}
\caption{Reviewed XGDAI techniques mapped to the four proposed phases of explanation.}
\label{fig_table_4phase}
\end{figure}

Fig. \ref{fig_table_4phase} shows how the reviewed XGDAI strategies in this work were mapped to the four proposed phases of explanation. As expected, a majority possessed elements for generating explanations while fewer have the ability to assess user comprehension and further fine-tune the explanation generation. Explanation communication however was a rarely explored domain. Meta-AQUA and Proactive were two methods that actively determined the communication channel as a viable technique to facilitate explanations. Future research directions may further explore the use of multimodal communication to improve the effectiveness of explainability methods.

\section{Conclusions and Further Work}

The field of XGDAI is emerging with many rich applications in several domains. Explainability enables transparency for these agents/robots and encourages users' trust for applications in safety-critical situations. This survey presents several techniques for explanation generation and communication proposed and implemented in XGDAI until today. Typically, many XGDAIs techniques can enable the robots/agents to justify and explain their decisions and plan rationales for their actions. However, the state of the art shows that current approaches on XGDAIs are still in their infancy, lacking an extensive explanation generation and communication framework. Consequently, this study highlights a framework towards a more extensive XGDAI that has an extended perceptual and cognitive explanation capability. Future work will involve developing a transparent and domain-agnostic integrated architecture for the effective actualization of explainability in XGDAI.

\section*{Acknowledgment}

This research was supported by the Georg Forster Research Fellowship for Experienced Researchers from Alexander von Humboldt-Stiftung/Foundation and Impact Oriented Interdisciplinary Research Grant (IIRG) (IIRG002C-19HWB) from University of Malaya. The authors gratefully acknowledge the support from the German Research Foundation (Deutsche Forschungsgemeinschaft / DFG) under project CML (TRR 169) and from the BMWK under project VeriKAS.



\bibliographystyle{IEEEtranN} 
\bibliography{draftarxiv.bib}

\newpage

\section*{Supplementary Materials}

\begin{center}
\begin{longtable}{|m{7.0cm}|m{1.6cm}|m{3.0cm}|m{1.2cm}|m{1.2cm}|}
\caption{Some taxonomy on XGDAI. Table presents highlights of key behavioral attributes and application domains of XGDAIs.} 
\label{tab1_supp} \\

\hline \multicolumn{1}{|m{7.0cm}|}{\textbf{Publication}} & \multicolumn{1}{|m{1.6cm}|}{\textbf{XGDAI Behavior}} & \multicolumn{1}{|m{3.0cm}|}{\textbf{Application Domain}} & \multicolumn{1}{|m{1.2cm}|}{\textbf{Trans- parency}} & \multicolumn{1}{|m{1.2cm}|}{\textbf{Domain specific}}\\ \hline 
\endfirsthead

\multicolumn{5}{c}%
{{\tablename\ \thetable{} -- continued from previous page}} \\
\hline \multicolumn{1}{|m{7.0cm}|}{\textbf{Publication}} & \multicolumn{1}{|m{1.6cm}|}{\textbf{XGDAI Behavior}} & \multicolumn{1}{|m{3.0cm}|}{\textbf{Application Domain}} & \multicolumn{1}{|m{1.2cm}|}{\textbf{Trans- parency}} & \multicolumn{1}{|m{1.2cm}|}{\textbf{Domain specific}}\\ \hline
\endhead

\hline \multicolumn{5}{|r|}{{Continued on next page}} \\ \hline
\endfoot

\hline 
\endlastfoot
 \cite{korpan2018toward}, \cite{jaidee2011integrated}, \cite{chen2014situation}, \cite{boyce2015effects}, \cite{chen2018situation}, \cite{borgo2018towards}, \cite{wicaksono2017towards}, \cite{baraka2016expressive}, \cite{kaptein2017personalised}, \cite{wortham2017improving}, \cite{takayama2011expressing}, \cite{pynadath2018clustering}, \cite{sridharan2019towards}, \cite{oudah2018ai}, \cite{hastie2018miriam}, \cite{gong2018behavior}, \cite{wohleber2017effects}, \cite{kaptein2017role}, \cite{wortham2016does}, \cite{rosenthal2016verbalization}, \cite{dragan2015effects}, \cite{novikova2015emotionally}, \cite{li2015communication},  \cite{van2013signaling}, \cite{lomas2012explaining}, \cite{jaidee2011case}, \cite{haynes2009designs}, \cite{kroske2009trusted}, \cite{yilmaz2019cognitive}, \cite{chang2020coactive}
 & Deliberative & Robot-human collaboration & Yes & Yes\\
 
 \hline
 \cite{chandrasekaran2017takes}, \cite{gervasio2018explanation}, \cite{chakraborti2017plan}, \cite{chakraborti2017visualizations}, \cite{lettl2013self}, \cite{klenk2013goal} 
 & Deliberative & Robot-human collaboration & - & No\\ 
 
 \hline
 \cite{chadalavada2015s}, \cite{gao2020joint} 
 & Reactive & Robot-human collaboration & Yes & No \\ 
 
 \hline
 \cite{hayes2017improving}, \cite{mikawa2018expression}, \cite{song2018effect}, \cite{floyd2016incorporating}, \cite{dannenhauer2018learning}, \cite{wang2016impact}, \cite{shindev2012exploration}, \cite{guzdial2018explainable}, \cite{groshev2018learning}, \cite{amir2018highlights}, \cite{ghosh2019generating}
 & Reactive & Robot-human collaboration & No & No \\ 
 
 \hline
 \cite{ginesi2020autonomous}, \cite{beckers2019intelligent}
 & Hybrid & Robot-human collaboration & No & Yes \\ 
 
 \hline
 \cite{sequeira2019interestingness}
 & Hybrid & Robot-human collaboration & No & No \\ 
 
 \hline
 \cite{korpan2018toward}, \cite{neerincx2018using} 
 & Hybrid & Robot-human collaboration & - & - \\ 
 
 \hline
 \cite{kroske2009trusted}, \cite{hofmarcher2019visual}, \cite{cultrera2020explaining}
 & Deliberative & Navigation & Yes & Yes \\
 
 \hline
 \cite{bourgais2019ben}
 & Deliberative & Navigation & Yes & No \\
	 
 \hline
 \cite{pan2020zero} 
 & Deliberative & Navigation & No & Yes \\
 
 \hline
 \cite{dannenhauer2018learning}, \cite{sukkerd2018toward}, \cite{jauffret2013self}, \cite{contreras2020unmanned}
 & Reactive & Navigation & No & No \\
 
 \hline
 \cite{qian2019deep}
 & Hybrid & Navigation & Yes & Yes \\
 
 \hline
 \cite{da2020agent}
 & Hybrid & Navigation & - & Yes \\
 
 \hline
 \cite{korpan2018toward}, \cite{wang2008robot}, \cite{santucci2016grail}
 & Hybrid & Navigation & - & No \\
 
 \hline
 \cite{bonaert2021explainable}
 & Hybrid & Navigation & No & No \\
 
 \hline
 \cite{munoz2010case} 
 & Deliberative & Game application & - & Yes \\
 
 \hline
 \cite{molineaux2018towards}, \cite{molineaux2010goal}, \cite{nau1999shop}, \cite{voelz1998rocco} 
 & Deliberative & Game application & - & No \\
 
 \hline
 \cite{ehsan2019automated}, \cite{ehsan2018rationalization}, \cite{amir2018highlights}, \cite{andreas2017translating}
 & Reactive & Game application & No & No \\
 
 \hline
 \cite{sahota1994reactive} 
 & Hybrid & Game application & - & No \\
 
 \hline
 \cite{sbai2013intra}, \cite{kroske2009trusted}
 & Deliberative & Search and Rescue & - & Yes \\
 
 \hline
 \cite{harbers2010design}, \cite{johnson1994agents}, \cite{broekens2010you}, \cite{harbers2009methodology},  \cite{kambhampati1994unified}, \cite{mooney1986domain} 
 & Deliberative & Training & Yes & Yes \\
 
 \hline
 \cite{huo2020hetropy}
 & Deliberative & Training & Yes & No \\
 
 \hline
 \cite{kaptein2017personalised}, \cite{kaptein2017role},  \cite{codella2018teaching}, \cite{grea2018explainable}
 & Deliberative & Healthcare & - & - \\
 
 \hline
 \cite{lauritsen2020explainable}, \cite{mirchi2020virtual}
 & Reactive & Healthcare & No & Yes \\
 
 \hline
 \cite{ginesi2020autonomous}
 & Hybrid & Healthcare & No & Yes \\
 
 \hline
 \cite{vermeulen2010improving}, \cite{lim2011design}, \cite{stumpf2010making}, \cite{rehman2005visually}, \cite{holliday2013effect}, \cite{tipaldi2020applying}, \cite{garcia2019human}, \cite{chatterjee2020xai4wind}
 & Reactive & Pervasive systems & No & Yes\\
 
 \hline
 \cite{li2020deep}
 & Hybrid & Pervasive systems & Yes & Yes \\
 
 \hline
 \cite{alkhabbas2020agent} 
 & Hybrid & Pervasive systems & No & No \\
 
 \hline
 \cite{xu2020knowledge} 
 & Hybrid & Recommender systems & No & No \\
 
 \hline
 \cite{holliday2013effect}, \cite{quijano2017make}, \cite{nilashi2016recommendation}, \cite{kulesza2013too}, \cite{kulesza2010explanatory} 
 & Reactive & Recommender systems & Yes & Yes \\
	 
 \hline
 \cite{mikawa2018expression} 
 & Reactive & Teleoperation & No & No \\
 
 \hline
 \cite{hastie2018miriam} 
 & Deliberative & Teleoperation & Yes & Yes \\
 
 \hline
 \cite{sbai2013intra}, \cite{hedhili2013explanation}, \cite{chang2020coactive}
 & Deliberative & Multi-agent systems & No & No \\
 
\end{longtable}
\end{center}

\begin{table}[htbp!]
\caption{Summary of explainability techniques for deliberative XGDAI}
\label{tab2_supp}
\begin{tabular}{|l|l|m{3cm}|m{2cm}|m{1.2cm}|m{1.2cm}|}
 
\hline \multicolumn{1}{|m{0.15\textwidth}|}{\textbf{Explainability Techniques}} & \multicolumn{1}{|m{0.23\textwidth}|}{\textbf{References}} & \textbf{What is explained?} & \textbf{Key triggers or basis for explanation generation} & \textbf{Trans-parent/ Post-hoc} & \textbf{Domain-specific/ Domain-agnostic} \\ \hline 

\multicolumn{1}{|m{0.15\textwidth}|}{Goal-driven Autonomy} & \multicolumn{1}{|m{0.23\textwidth}|}{\cite{molineaux2010goal}, \cite{weber2012learning}, \cite{floyd2016incorporating}} & Plans, goals & Expectation failures, states mismatched & T & DS \\ \hline
 
\multicolumn{1}{|m{0.15\textwidth}|}{Explainable BDI} & \multicolumn{1}{|m{0.23\textwidth}|}{\cite{harbers2010design}, \cite{flin2017incident}} & Actions & Underlying belief, desire, and intention & T & DS \\ \hline
 
\multicolumn{1}{|m{0.15\textwidth}|}{SAT} & \multicolumn{1}{|m{0.23\textwidth}|}{\cite{endsley2018innovative}} & Agent's current state, goals, plans, reasoning process, and future projections & Underlying beliefs and goals & T & DS \\ \hline
 
\multicolumn{1}{|m{0.15\textwidth}|}{Meta-AQUA} & \multicolumn{1}{|m{0.23\textwidth}|}{\cite{cox2007perpetual}} & Unusual events or states of the world, agent's reasoning & Anomaly (states mismatched) or interesting event & T & DA \\ \hline
 
\multicolumn{1}{|m{0.15\textwidth}|}{Proactive Explanation} & \multicolumn{1}{|m{0.23\textwidth}|}{\cite{gervasio2018explanation}} & Agent's decisions & Anticipated surprise or potential expectation violation. & T & DA \\ \hline
 
\multicolumn{1}{|m{0.15\textwidth}|}{KAGR} & \multicolumn{1}{|m{0.23\textwidth}|}{\cite{sbai2013intra}} & Agent's reasoning & Events & P & DA \\ \hline
 
\multicolumn{1}{|m{0.15\textwidth}|}{eXplainable-Plan} & \multicolumn{1}{|m{0.23\textwidth}|}{\cite{korpan2018toward}, \cite{bidot2010verbal}, \cite{borgo2018towards}} & Agent's plan & Plan mismatched & P/T & DA \\ \hline
 
\multicolumn{1}{|m{0.15\textwidth}|}{Explainable NPC} & \multicolumn{1}{|m{0.23\textwidth}|}{\cite{molineaux2018towards}, \cite{van2004explainable}} & Agent's actions & On supervisor's request or after active goal completion & P & DA \\ \hline
 
\multicolumn{1}{|m{0.15\textwidth}|}{Debrief} & \multicolumn{1}{|m{0.23\textwidth}|}{\cite{johnson1994agents}} & Agent's decisions & On request or after active goal completion & T & DS \\ \hline
 
\multicolumn{1}{|m{0.15\textwidth}|}{Explainable CREATIVE Model} & \multicolumn{1}{|m{0.23\textwidth}|}{\cite{wicaksono2017towards}, \cite{lavrac1994inductive}} & Agent's actions (tool creation process) & On request & T & DS \\ \hline
\end{tabular}
\end{table}

\begin{table}[htbp!]
 \centering
 \caption{Summary of explainability techniques for reactive and hybrid XGDAIs}
 \label{tab3_supp}
 \begin{tabular}{|l|l|m{2.2cm}|m{2cm}|m{1.3cm}|m{1cm}|m{1cm}|}
 \hline
 \multicolumn{1}{|m{0.15\textwidth}|}{Explainability Techniques} & References & What is explained? &
 Key triggers or basis for explanation generation & Reactive/ Hybrid agents & Trans-parent/ Post-hoc & Domain-specific/ Domain-agnostic \\ \hline
 
 \multicolumn{1}{|m{0.15\textwidth}|}{ARG} & \multicolumn{1}{|m{0.15\textwidth}|}{\cite{ehsan2019automated}} & Agent's internal state and action & After each action completes & R & P & DA \\ \hline
 
 \multicolumn{1}{|m{0.15\textwidth}|}{XRL} & \multicolumn{1}{|m{0.15\textwidth}|}{\cite{cruz2019memory}, \cite{juozapaitis2019explainable}, \cite{tabrez2019improving}, \cite{littman1994memoryless}, \cite{sequeira2019interestingness}} & Agent's decisions, actions or behavior & After active goal completes, perceived user-robot model disparity & R & P & DA \\ \hline
 
 \multicolumn{1}{|m{0.15\textwidth}|}{APE} & \multicolumn{1}{|m{0.15\textwidth}|}{\cite{hayes2017improving}} & Control policies & Inquiry by users & R & P/T & DA \\ \hline
 
 \multicolumn{1}{|m{0.15\textwidth}|}{PeCoX} & \multicolumn{1}{|m{0.15\textwidth}|}{\cite{neerincx2018using}, \cite{dennett1978three}, \cite{harbers2010guidelines} \cite{doring2003explaining}} & Agent's behavior & Request of the user & H & P & DA \\ \hline
 \end{tabular}

\end{table}

\begin{table}[htbp!]
\centering
\caption{Summary of explanation communication techniques for XGDAIs.}
\label{tab4_supp}
\begin{tabular}{|m{0.15\textwidth}|m{0.25\textwidth}|m{0.5\textwidth}|}
\hline
Explanation Communication Types & Techniques & References\\ \hline

\multirow{4}{0.15\textwidth}{Visualization} & 

CAM (NN Approach) & 
\cite{zhou2016learning}, 
\cite{selvaraju2017grad},
\cite{ng2018classification}, 
\cite{kerzel2022whats},
\cite{siqueira2020efficient}, 
\cite{lucey2010extended} \\ \cline{2-3} & 

Visual grounding via structured graph (NS approach) & 
\cite{conser2019revisiting}, 
\cite{shridhar2020ingress}, 
\cite{johnson2015image}, 
\cite{xu2017scene}, 
\cite{ren2016faster}, 
\cite{yang2018graph},
\cite{zhou2018weakly} \\ \cline{2-3} & 

Automata (NS approach) & 
 \cite{giles1992learning},
 \cite{zeng1993learning},
 \cite{omlin1996extraction},
 \cite{weiss2018extracting},
 \cite{wermter1999preference},
 \cite{arevian2003symbolic},
 \cite{wermter2000knowledge}\\ \cline{2-3} & 
 
Rules Extraction Algorithms (NS approach) & 
 \cite{andrews1995survey} \\ \hline
 
\multirow{2}{0.15\textwidth}{Numerical Form} & 

Attention Weights & 
 \cite{lipton2015learning},
 \cite{choi2016retain}, 
 \cite{sha2017interpretable},
 \cite{bai2018interpretable},
 \cite{lee2017interactive},
 \cite{liu2018visual} \\ \cline{2-3} & 
 
SHapley Additive exPlanation & 
 \cite{lundberg2017unified} \\ \hline

\multirow{2}{0.15\textwidth}{Textual Communication} & 
 
Textual Data Logs & 
 \cite{harbers2010design},
 \cite{van2004explainable} \\ \cline{2-3} & 
 
Prolog & 
 \cite{wicaksono2017towards} \\ \hline
 
\multirow{2}{0.15\textwidth}{Verbal Communication} & 
 
Verbal Communication & 
 \cite{kerzel2022whats},
 \cite{breazeal2008social},
 \cite{ambsdorf2022explain},
 \cite{shirai1973ear},
 \cite{shirai1974algorithm},
 \cite{kobayashi1985speech},
 \cite{osada2006scenario},
 \cite{sato2006auditory},
 \cite{sakagami2002intelligent},
 \cite{nisimura2002aska},
 \cite{sabelli2011conversational} \\ \cline{2-3} &
  
Referring Expression & 
 \cite{shridhar2020ingress},
 \cite{yu2017a},
 \cite{foster2008the},
 \cite{fang2013towards},
 \cite{giuliani2010situated},
 \cite{dale1995computational},
 \cite{kranstedt2005incremental} \\ \hline

\multirow{2}{0.15\textwidth}{Non-verbal Cues} & 
 
Lights & 
 \cite{kerzel2022whats},
 \cite{baraka2016expressive},
 \cite{song2018effect} \\\cline{2-3} & 
 
Expressive motion & 
 \cite{mikawa2018expression}, 
 \cite{admoni2017social} \\ \hline

\end{tabular}
\end{table}

\clearpage

\end{document}